%% file: acl_latex.tex
\pdfoutput=1

\documentclass[11pt]{article}

\usepackage[final]{acl}

\usepackage{times}
\usepackage{latexsym}
\usepackage[T1]{fontenc}
\usepackage[utf8]{inputenc}
\usepackage{microtype}
\usepackage{inconsolata}
\usepackage{graphicx}
\usepackage{microtype}
\usepackage{tabularx}
\usepackage[table]{xcolor}

\usepackage{amsmath}
\usepackage{amssymb}
\usepackage{mathtools}
\usepackage{amsthm}
\usepackage{multirow}
\usepackage{makecell}
\usepackage{booktabs}
\usepackage{array}
\usepackage{longtable}
\usepackage{subcaption}
\usepackage{fontawesome}
\usepackage{centernot}
\usepackage{pgf} 
\usepackage{tikz} 
\usepackage{stfloats}

\theoremstyle{plain}

\theoremstyle{definition}

\theoremstyle{remark}

\usepackage[most,skins,theorems]{tcolorbox}
\tcbset{
  aibox/.style={
    width=\linewidth,
    top=8pt,
    bottom=4pt,
    colback=blue!6!white,
    colframe=black,
    colbacktitle=black,
    enhanced,
    center,
    attach boxed title to top left={yshift=-0.1in,xshift=0.15in},
    boxed title style={boxrule=0pt,colframe=white,},
  }
}
\newtcolorbox{AIbox}[2][]{aibox,title=#2,#1}

\tcbset{
    promptstyle/.style={
        enhanced,
        colback=white,
        colframe=black,
        colbacktitle=gray!20,
        width=0.98\linewidth, 
        coltitle=black,
        rounded corners,
        sharp corners=north,
        boxrule=0.5pt,
        drop shadow=black!50!white,
        attach boxed title to top left={
            xshift=-2mm,
            yshift=-2mm
        },
        title code={%
            \begin{minipage}{0.8\linewidth}
                \centering\thetitle
            \end{minipage}
        },
        boxed title style={
            rounded corners,
            size=small,
            colback=gray!20,
        },
        fonttitle=\normalsize\bfseries,
        before upper={\parindent15pt}
    }
}

\usepackage{ifthen} 
\newcommand{\coloredAvg}[1]{%
  \pgfmathsetmacro{\val}{#1}%
  \pgfmathsetmacro{\absval}{abs(\val)}%
  \pgfmathsetmacro{\perc}{min(100,100*\absval/3)}%
  \ifthenelse{\lengthtest{\val pt > 0pt}}%
    {\cellcolor{red!\perc}#1}%
    {\ifthenelse{\lengthtest{\val pt < 0pt}}%
      {\cellcolor{green!\perc}#1}%
      {\cellcolor{white}#1}}%
}

\tcbset{
    userstyle/.style={
        enhanced,
        colback=white,
        colframe=black,
        colbacktitle=gray!20,
        coltitle=black,
        rounded corners,
        sharp corners=north,
        boxrule=0.5pt,
        drop shadow=black!50!white,
        attach boxed title to top left={
            xshift=-2mm,
            yshift=-2mm
        },
        boxed title style={
            rounded corners,
            size=small,
            colback=gray!20
        }
    },
    replystyleg/.style={
        enhanced,
        colback=green!15,
        colframe=black,
        colbacktitle=green!30,
        coltitle=black,
        boxrule=0.5pt,
        drop shadow=black!50!white,
        rounded corners,
        sharp corners=north,
        attach boxed title to top right={
            xshift=-2mm,
            yshift=-2mm
        },
        boxed title style={
            rounded corners,
            size=small,
            colback=green!40
        }
    },
    replystyler/.style={
        enhanced,
        colback=red!15,
        colframe=black,
        colbacktitle=red!40,
        coltitle=black,
        boxrule=0.5pt,
        drop shadow=black!50!white,
        rounded corners,
        sharp corners=north,
        attach boxed title to top right={
            xshift=-2mm,
            yshift=-2mm
        },
        boxed title style={
            rounded corners,
            size=small,
            colback=red!40
        }
    }
}

\newtcolorbox{prompt}[2][]{
    colback=white,
    colframe=gray!45,
    fonttitle=\bfseries,
    coltitle=black,
    sharp corners,
    title=#2,
    #1
}

\tcbset{
    promptstyle/.style={
        enhanced,
        colback=white,
        colframe=black,
        colbacktitle=gray!20,
        coltitle=black,
        rounded corners,
        sharp corners=north,
        boxrule=0.5pt,
        drop shadow=black!50!white,
        attach boxed title to top left={
            xshift=-2mm,
            yshift=-2mm
        },
        boxed title style={
            rounded corners,
            size=small,
            colback=gray!20
        }
    },
    replystyleg/.style={
        enhanced,
        colback=green!15,
        colframe=black,
        colbacktitle=green!30,
        coltitle=black,
        boxrule=0.5pt,
        drop shadow=black!50!white,
        rounded corners,
        sharp corners=north,
        attach boxed title to top right={
            xshift=-2mm,
            yshift=-2mm
        },
        boxed title style={
            rounded corners,
            size=small,
            colback=green!40
        }
    },
    replystyler/.style={
        enhanced,
        colback=red!15,
        colframe=black,
        colbacktitle=red!40,
        coltitle=black,
        boxrule=0.5pt,
        drop shadow=black!50!white,
        rounded corners,
        sharp corners=north,
        attach boxed title to top right={
            xshift=-2mm,
            yshift=-2mm
        },
        boxed title style={
            rounded corners,
            size=small,
            colback=red!40
        }
    }
}

\newtcolorbox{promptbox}[1][]{
    promptstyle,
    title=Prompt,
    #1
}

\title{Small Models Struggle to Learn from Strong Reasoners}

\author{
\textbf{Yuetai Li}\textsuperscript{$\clubsuit$} \;\;\;  
\textbf{Xiang Yue}\textsuperscript{$\diamondsuit$} \;\;\; 
\textbf{Zhangchen Xu}\textsuperscript{$\clubsuit$} \;\;\;  
\textbf{Fengqing Jiang}\textsuperscript{$\clubsuit$} \;\;\;
\textbf{Luyao Niu}\textsuperscript{$\clubsuit$} \;\;\; \\ 
\textbf{Bill Yuchen Lin}\textsuperscript{$\clubsuit$} \;\;\;
\textbf{Bhaskar Ramasubramanian}\textsuperscript{$\spadesuit$} \; \;\;
\textbf{Radha Poovendran}\textsuperscript{$\clubsuit$}\\
  \textsuperscript{$\clubsuit$}University of Washington \; 
  \textsuperscript{$\diamondsuit$}Carnegie Mellon University \;
  \textsuperscript{$\spadesuit$}Western Washington University \\
  \texttt{\{yuetaili,zxu9,fqjiang,luyaoniu,byuchenl,rp3\}@uw.edu},\\
  \texttt{xyue2@andrew.cmu.edu}, \texttt{ramasub@wwu.edu} \vspace{1em} \\
   \textbf{Huggingface}: \url{https://huggingface.co/UWNSL} \\
   \textbf{Project Page}: \url{https://small-model-gap.github.io/}   
}

\begin{document}
\maketitle
\begin{abstract}

Large language models (LLMs) excel in complex reasoning tasks, and distilling their reasoning capabilities into smaller models has shown promise. However, we uncover an interesting phenomenon, which we term the \textit{Small Model Learnability Gap}: small models ($\leq$3B parameters) do not consistently benefit from long chain-of-thought (CoT) reasoning or distillation from larger models. Instead, they perform better when fine-tuned on shorter, simpler reasoning chains that better align with their intrinsic learning capacity. To address this, we propose Mix Distillation, a simple yet effective strategy that balances reasoning complexity by combining long and short CoT examples or reasoning from both larger and smaller models. Our experiments demonstrate that Mix Distillation significantly improves small model reasoning performance compared to training on either data alone. These findings highlight the limitations of direct strong model distillation and underscore the importance of adapting reasoning complexity for effective reasoning capability transfer.

\end{abstract}

\input{sec/intro_v2}

\input{sec/preliminaries}

\input{sec/empirical}

\input{appendix/related}

\input{sec/conclusion}

\input{sec/limitation}
\input{sec/ethical}

\bibliography{custom}

\clearpage
\appendix
\input{appendix/more_setup}

\input{appendix/more_exp_data}
\input{appendix/examples_of_different_CoT}

\end{document}

%% file: sec/intro_v2.tex
\section{Introduction}

Large language models (LLMs) \citep{anthropic2023claude, brown2020languagemodelsfewshotlearners, openai2023gpt4,touvron2023llamaopenefficientfoundation} have demonstrated remarkable performance in complex reasoning tasks, enabling advancements in mathematical problem-solving, logical inference, and structured decision-making \citep{cobbe2021trainingverifierssolvemath, shao2024deepseekmathpushinglimitsmathematical, yang2024qwen25mathtechnicalreportmathematical}. A key advancement in
improving LLM complex reasoning capability is the chain-of-thought (CoT) prompting. This technique decomposes complex problems into intermediate reasoning steps, enhancing both performance and interpretability. \citep{wei2023chainofthoughtpromptingelicitsreasoning}.

\begin{figure}[h]
    \centering
    \includegraphics[width=\linewidth]{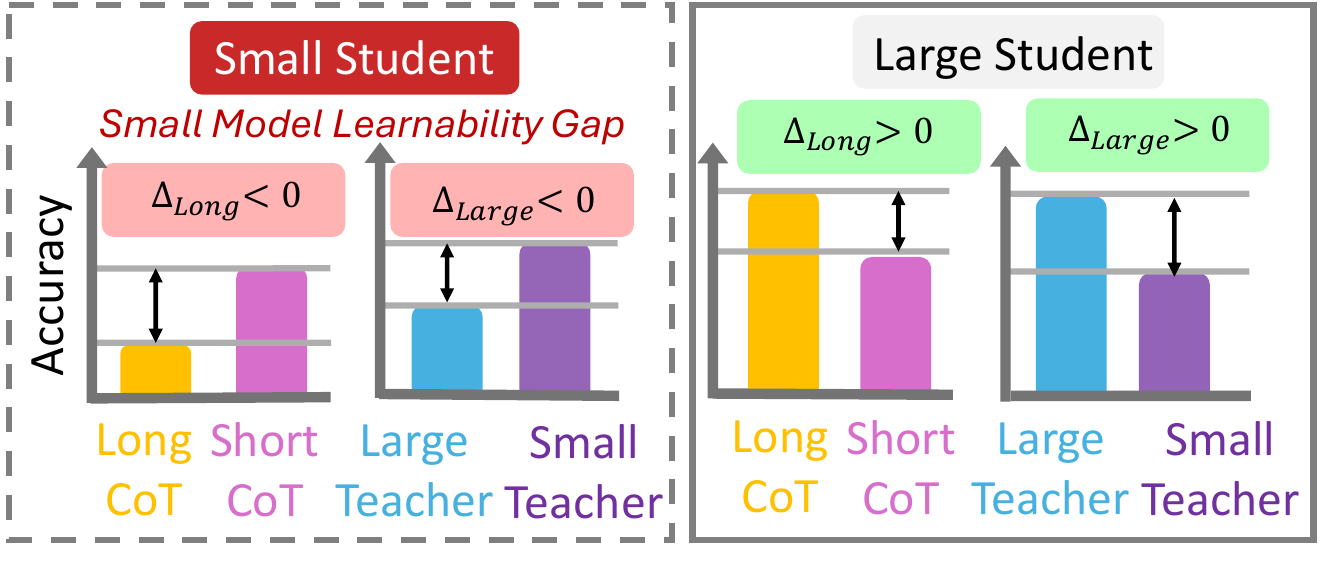}

    \caption{Small student models ($\leq$3B parameters) do not consistently benefit from long CoT reasoning or distillation from large teacher models. Instead, they perform better when fine-tuned on shorter CoT reasoning or distilled from smaller teachers, which better matches their intrinsic learning capacity. We term this phenomenon the \textit{Small Model Learnability Gap}.}
    \label{fig:main}
    \vspace{-1em}
\end{figure}

However, the high computational cost of LLMs hinders their deployment on resource-constrained devices, motivating the development of smaller models that offer similar capabilities at reduced cost. A widely adopted strategy to achieve this is distillation \citep{agarwal2024onpolicydistillationlanguagemodels, hinton2015distillingknowledgeneuralnetwork, kim2024promptkddistillingstudentfriendlyknowledge}, where CoT sequences generated by a strong teacher model are used to fine-tune a weaker student model.
Naturally, one might expect that distilling CoT sequences from stronger models would consistently improve small models' complex reasoning capabilities \citep{agarwal2024onpolicydistillationlanguagemodels, deepseekai2024deepseekv3technicalreport, min2024imitateexploreselfimprovereproduction, tunstall2023zephyrdirectdistillationlm}.

However, we reveal an interesting phenomenon, which we term the \textit{Small Model Learnability Gap} (Fig. \ref{fig:main}): small models do not consistently benefit from the complex reasoning sequences provided by strong teachers, such as long CoT reasoning or distillation from large models. In our experiments, we observe that when small models are exposed to long and intricate reasoning traces, they struggle to internalize the multi-step logic due to their constrained ability. Instead, small models perform better when fine-tuned on \textit{shorter, simpler reasoning chains} that align more closely with their intrinsic learning capacity. This suggests that small models struggle to process overly elaborate reasoning traces or adapt to the distribution shifts introduced by stronger teachers, ultimately limiting their ability to generalize effectively.

To address the challenge described above, we propose \textit{Mix Distillation}, a simple yet effective approach that balances reasoning complexity by blending different types of reasoning traces. Specifically, our method comprises two configurations: (1) \textit{Mix-Long} – A combination of long and short CoT examples, ensuring that small models are exposed to both detailed and concise reasoning steps. (2) \textit{Mix-Large} – A mixture of responses from both larger and smaller models, allowing small models to learn from reasoning chains that are better suited to their capacity.  

Our experiments demonstrate that \textit{Mix Distillation} consistently improves small model reasoning performance compared to standard distillation. 
For instance, \texttt{Qwen2.5-3B-Instruct} improves by more than 8 points on MATH and AMC using Mix-Long, compared to direct training on long CoT data.
\texttt{Qwen2.5-3B-Instruct} gains more than 7 points on MATH, AIME and AMC using Mix-Large compared with training on large teacher CoT data.

These findings highlight a fundamental limitation of direct strong model distillation and emphasize the importance of \textit{adapting reasoning complexity} for effective knowledge transfer. By carefully designing fine-tuning strategies, we provide new insights into overcoming the constraints of small model learning, making them more effective at reasoning-intensive tasks.

%% file: sec/preliminaries.tex
\section{Preliminaries}

\subsection{Notation}
Let $x = (x_1, x_2, \dots, x_n)$ represent an input sequence (e.g., a prompt), and $y = (y_1, y_2, \dots, y_m)$ be the corresponding output sequence.
We consider a LLM parameterized by $\theta$, which predicts the next token following a conditional distribution 
$\pi_\theta \bigl(y_t|x, y_{1:t-1}\bigr)$. We denote by $\text{CoT}(y) \subseteq y$ the subset of tokens in the generated output that encodes a \emph{chain-of-thought}, often serving as a reasoning trace or explanatory sequence.

Throughout this work, we use the term \textbf{short CoT}, to describe concise reasoning paths to arrive at solutions \citep{min2024imitateexploreselfimprovereproduction,yeo2025demystifyinglongchainofthoughtreasoning} and \textbf{long CoT} to describe an extended reasoning sequence that is not only longer but also demonstrates more complex reflective thoughts \citep{QwenTeam2024b, yeo2025demystifyinglongchainofthoughtreasoning}. Additionally, we use the term \textbf{large teacher CoT} to refer to the reasoning trace generated by a larger teacher model, and the term \textbf{small teacher CoT} for the reasoning steps produced by a smaller teacher model.
Please see Appendix \ref{app:example} for more examples.

\subsection{Supervised Fine-Tuning (SFT)} 

Supervised fine-tuning (SFT) is widely adopted to enhance reasoning capabilities of LLMs on a dataset $\mathcal{D} = \{(x^i, y^i)\}_{i=1}^N$, where $y^i$ can be short CoT, long CoT, strong model CoT or weak model CoT sequences.
The SFT process updates the parameters $\theta$ of a language model by minimization the negative log-likelihood loss over the instruction dataset $\mathcal{D}$.

%% file: sec/empirical.tex
\section{Small Model Learnability Gap}
\label{sec: empirical}
In this section, we fine-tune student models using different CoT data. 
We then reveal the small model learnability gap given the performance of fine-tuned models.

\subsection{Experiment Setup}

\paragraph{Datasets.}
We use the 7,500 prompt set of MATH \citep{hendrycks2021measuringmathematicalproblemsolving}. This dataset encompasses seven math topics such as advanced calculus, geometry, and linear algebra.

\paragraph{Student models.}
Our study considers ten student models from the Qwen \citep{qwen2.5} and Llama \citep{llama32,llama31} model families of varying sizes. These models include the Instruct version of \texttt{Qwen2.5-0.5B}, \texttt{Qwen2.5-1.5B}, \texttt{Qwen2.5-3B}, \texttt{Qwen2.5-7B}, \texttt{Qwen2.5-14B}, and \texttt{Qwen2.5-32B}, and the Instruct version of \texttt{Llama3.2-1B}, \texttt{Llama3.2-3B}, \texttt{Llama3.1-8B}, and \texttt{Llama3.3-70B}. A comprehensive overview of the student models is presented in Table \ref{tab:models_overview} of Appendix \ref{appendix:More on Experimental Setups}.

\paragraph{Teacher models.}

To compare long CoT with short CoT, we use \texttt{QwQ-32B-Preview} \citep{QwenTeam2024b} to generate long CoT sequences and \texttt{Qwen2.5-32B-Instruct} as the response generator for short CoT. 
Within each model family, we designate the larger scale model as the large teacher and the smaller scale model as the small teacher. 
This includes \texttt{Qwen2.5-72B-Instruct} vs \texttt{Qwen2.5-\allowdisplaybreaks3B-Instruct},~\texttt{Llama3.1-70B-Instruct} vs~\texttt{Llama3.1-8B-Instruct},~and \texttt{Gemma2-27B-it} vs \texttt{Gemma2-9B-it}.

\begin{figure*}[!t]
    \centering
    \includegraphics[width=\textwidth]{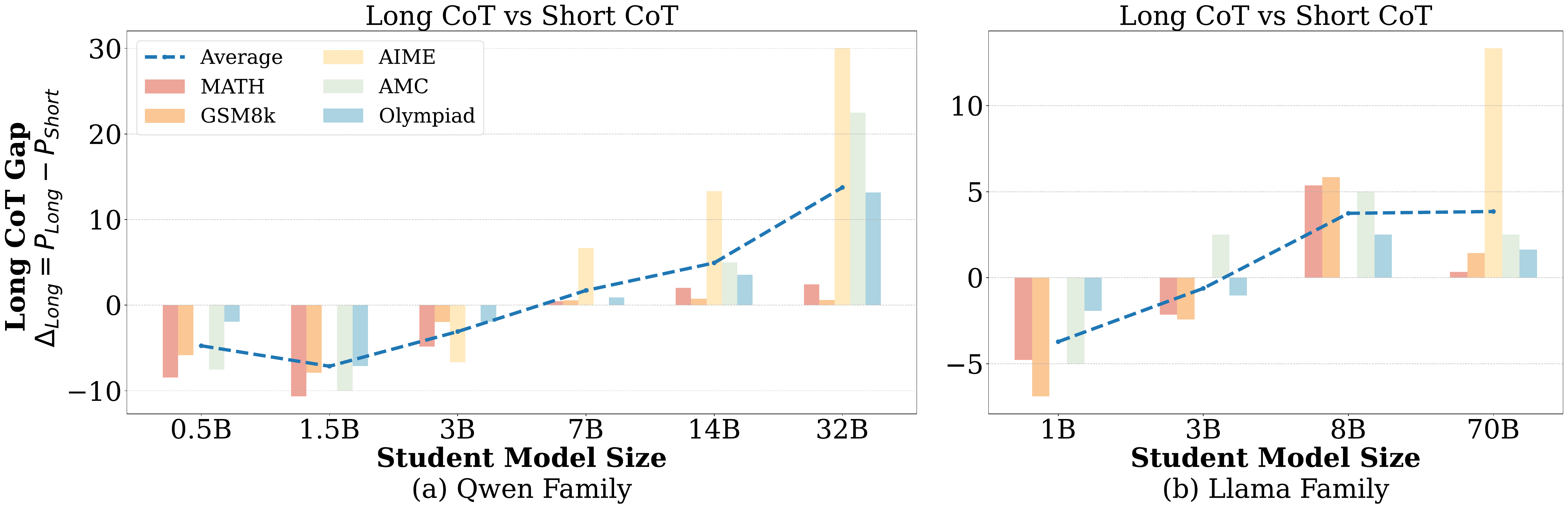}
    \caption{Long CoT Gap ($\Delta_{Long}=P_{Long} - P_{Short}$) of student models with different models sizes for (a) Qwen family (b) Llama family. For teacher models, \texttt{QwQ-preview-32B} is chosen to generate long CoT responses, while \texttt{Qwen2.5-32B-Instruct} is chosen to generate short CoT responses. Negative (Positive) $\Delta_{Long}$ indicates that long CoT is worse (better) than short CoT. Our results demonstrate that short CoT is better for smaller student models (indicated by $\Delta_{Long}$ < 0), while long CoT is better for larger student models (indicated by $\Delta_{Long}$ > 0).}
    \label{fig:combined_model_performance}
\end{figure*}
\begin{figure*}[!t]
    \centering
    \includegraphics[width=\textwidth]{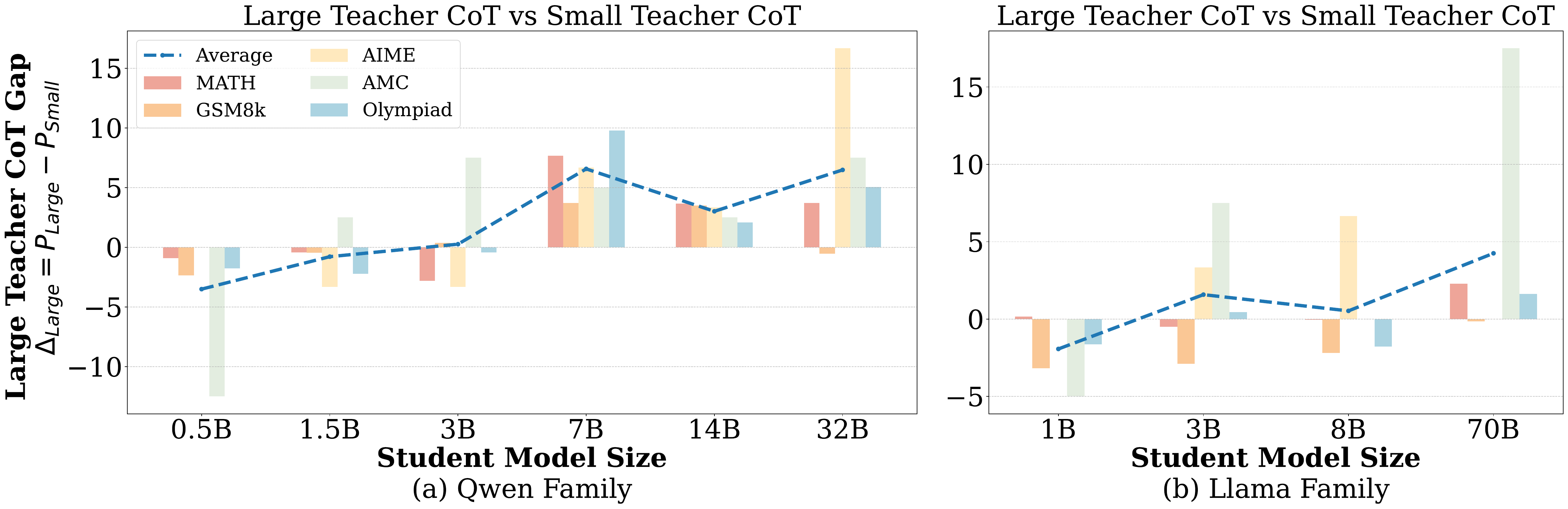}
    \caption{Large model CoT Gap ($\Delta_{Large}=P_{Large} - P_{Small}$) of student models with different models sizes for (a) Qwen family (b) Llama family. For teacher models, \texttt{Qwen2.5-72B-Instruct} is chosen as the large teacher to generate responses, while \texttt{Qwen2.5-3B-Instruct} is chosen as the small teacher to generate responses. Negative (positive) $\Delta_{Large}$ indicates that large teacher CoT is worse (better) than small teacher CoT. Our results demonstrate that small teacher CoT is better for smaller student models (indicated by $\Delta_{Large}$ < 0), while large model CoT is better for larger student models (indicated by $\Delta_{Large}$ > 0).}
    \label{fig:lmp_combined_model_performance}
\end{figure*}

\paragraph{Evaluation Benchmarks.}

We evaluate the reasoning capability of fine-tuned student models on a set of commonly used benchmarks, including MATH \citep{hendrycks2021measuringmathematicalproblemsolving}, GSM8K \citep{cobbe2021trainingverifierssolvemath}, AMC 2023, AIME 2024, and the English math subset of OlympiadBench \citep{he2024olympiadbenchchallengingbenchmarkpromoting}. These benchmarks span a wide range of challenge levels, from elementary mathematics to advanced competition problems. 
We define the student model performance as the average score on five benchmarks. Unless otherwise specified, all fine-tuned models are evaluated in a zero-shot setting using greedy decoding. We set the maximum generation tokens as 16k. Please see Appendix \ref{appendix:More on Experimental Setups} for detailed experimental setup.

We define the following performance scores:
\begin{itemize}
    \item \(P_{Long}\): Performance score of a student model fine-tuned on long CoT data.
    \item \(P_{Short}\): Performance score of a student model fine-tuned on short CoT data.
    \item \(P_{Large}\): Performance score of a student model fine-tuned on CoT from a larger teacher.
    \item \(P_{Small}\): Performance score of a student model fine-tuned on CoT from a smaller teacher.
\end{itemize}

\paragraph{Training Setup.} 
Teacher models generate responses by rejection sampling 
\citep{dong2023raftrewardrankedfinetuning, gulcehre2023reinforcedselftrainingrestlanguage, tong2024dartmathdifficultyawarerejectiontuning, yuan2023scalingrelationshiplearningmathematical, yue2023mammothbuildingmathgeneralist, zelikman2022starbootstrappingreasoningreasoning}
By default, teacher models employ greedy decoding.  
By combining the math problem instructions with corresponding solutions generated by teacher models, we construct problem-solution pairs to fine-tune student models. 
We train the models using the LLaMA-Factory framework \citep{zheng2024llamafactory}. 
For student models of scale less than 14B, we use full-parameter SFT and implement a cosine learning rate schedule with a maximum learning rate of $10^{-5}$ to fine-tune student models for two epochs \citep{touvron2023llama}. 
For student models larger than 14B, we adopt LoRA fine-tuning with a learning rate of $10^{-4}$ for two epochs. Detailed hyperparameters and information about the experimental platform are provided in Appendix \ref{appendix:More on Experimental Setups}.

\subsection{Long CoT Gap}
This section evaluates the reasoning capabilities of student models fine-tuned over long CoT data and short CoT data. We quantify the performance difference between long and short CoT data using \emph{long CoT gap} \(\Delta_{Long}\), defined as:
\begin{equation*}
\Delta_{Long} = P_{Long} - P_{Short}.
\end{equation*}

Figure \ref{fig:combined_model_performance} provides a comprehensive overview of the long CoT gap $\Delta_{Long}$ across different student models. 
The detailed benchmark scores on MATH, GSM8K, AIME, AMC, and OlympiadBench are deferred to Table \ref{tab:full_performance_lg} in Appendix \ref{appendix: More Experiments}.
We report the following key takeaways. 

\begin{AIbox}{Takeaway 1: Long CoT Gap}
Small student models tend to benefit more from short CoT, while large student models gain greater advantages from long CoT.
\end{AIbox}

We observe that long CoT is more effective for larger models, consistently leading to improved performance across most math benchmarks. 
For example, the student model \texttt{Qwen2.5-32B-Instruct} improves about 15 points across all math metrics on average. 

However, long CoT data is not effective for smaller models, yielding significantly less improvement compared to short CoT. On the MATH and AMC benchmarks, student model \texttt{Qwen2.5-1.5B-Instruct}  performs over 10 points lower when fine-tuned with long CoT data. This shows that smaller models may not be able to effectively learn and utilize the long CoT paradigm. 
Please see more attribution analysis in Section \ref{More Analysis Results}.

\input{tables/long_cot_compare}
\input{tables/lmp_compare}

\subsection{Large Teacher CoT Gap}
We investigate how effective small models may learn from large teacher and small teachers.
We define a \emph{large teacher CoT gap} as:
\[
\Delta_{Large} = P_{Large} - P_{Small}.
\]

Figure \ref{fig:lmp_combined_model_performance} provides a comprehensive comparison of the $\Delta_{Large}$ incurred by all student models. 
The detailed benchmark scores of MATH, GSM8K, AIME, AMC and OlympiadBench are deferred to Table \ref{tab:lmp-full_comparison} in Appendix \ref{appendix: More Experiments}. More experimental results of different teacher models, including \texttt{Llama3.1-70B} vs \texttt{Llama3.1-8B} and \texttt{Gemma2-27B} vs \texttt{Gemma2-9B} are in Table \ref{tab:lmp_comparison2} of Appendix \ref{appendix: More Experiments}.

We observe that larger student models learn effectively from large teacher CoT. 
For example, \texttt{Qwen2.5-7B-Instruct} and \texttt{Qwen2.5-32B-Instruct} student models improve over 5 points on average, with \texttt{Qwen2.5-32B-Instruct} achieving more than a 15 point increase on the AIMC benchmark. 
However, smaller models do not learn effectively from large teacher models such as \texttt{Qwen2.5-72B-Instruct}. 
Instead, small teacher models such as \texttt{Qwen2.5-3B-Instruct} may serve as better teacher models for small student models.
For instance, the performance of \texttt{Qwen2.5-0.5B-Instruct} degrades by more than 10 points on the AMC benchmark.

We remark that both larger teachers and small teachers generate short CoT data in this section to fine-tune student models, with no significant difference in average length. Specifically, the average token length is 432.98 for the 72B teacher and 440.70 for the 3B teacher. This helps eliminating CoT length as a confounding variable in our results of the large teacher CoT gap.

Note that prior studies \citep{kim2024evaluatinglanguagemodelssynthetic} also demonstrated that stronger models are not necessarily stronger teachers, 
emphasizing response generator and teacher-side factors. Our work differs in that we attribute this phenomenon primarily to the size of the student model.

\begin{AIbox}{\makecell{Takeaway 2: Large Teacher CoT Gap}}
Small student models tend to learn better from small teachers, while large student models benefit more from large teachers.
\end{AIbox}

\input{tables/mix_distilation_qwen}

\subsection{Analysis of Small Model Learnability Gap}
\label{More Analysis Results}

\paragraph{Domain knowledge affects learnability gap.}

\begin{figure*}[!t]
    \centering
    \includegraphics[width=\textwidth]{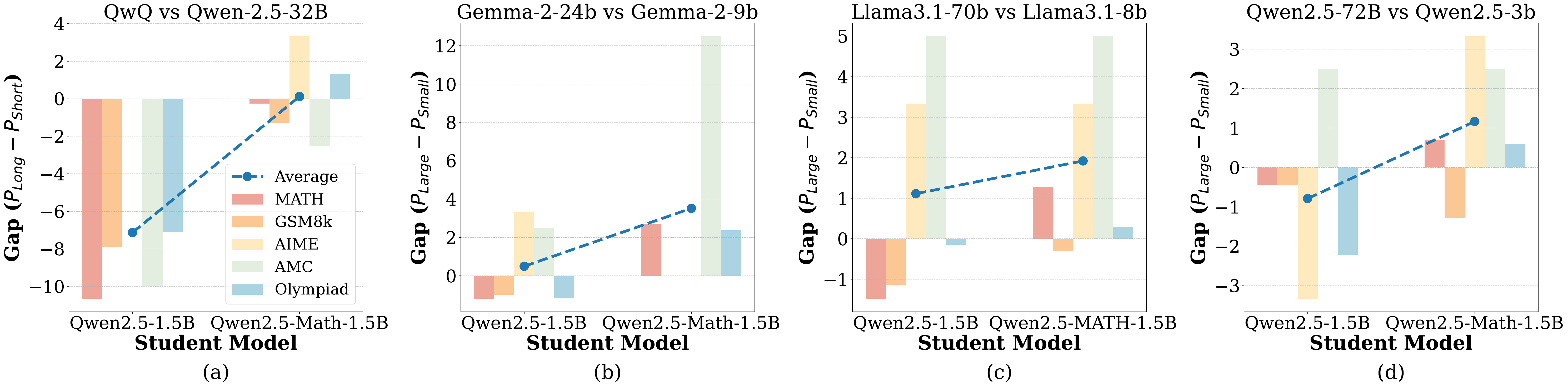}
    \caption{Math expert models usually have a less significant Learnability Gap than the general models. 
    A positive Gap means long CoT or large teacher CoT is better while negative means worse. This indicates that the math expert model could more easily learn from long CoT data or large teacher CoT. }
    \label{fig:math_expert_vs_general}
\end{figure*}

We observe that math expert models, in spite of small model size,  exhibit a smaller learnability gap for both long CoT and large teacher CoT data compared to general models in Figure \ref{fig:math_expert_vs_general}. 
Specifically, we compare the learnability gaps between the student models \texttt{Qwen2.5-Math-1.5B-Instruct} and \texttt{Qwen2.5-1.5B-Instruct}. Our findings show that the long CoT gap of the small math expert model is significantly smaller than that of general small models.
Furthermore, the performance improvement of \texttt{Qwen2.5-Math-1.5B} when fined-tuned with large teacher CoT exceeds that of \texttt{Qwen2.5-1.5B}, suggesting that math expert models benefit more substantially from large teacher CoT. We conjecture that a key factor leading to the small model learnability gap is the \textit{limited in-domain knowledge of small student models}.
We summarize this observation in the following takeaway.

\begin{AIbox}{\makecell{Takeaway 3: Domain Knowledge Helps}} Expert models with mid-training usually exhibit a less significant Learnability Gap compared to general models.  \end{AIbox}

\paragraph{Distribution Mismatch between student and teacher models.} One contributing factor to Small Model Learnability Gap is that small models struggle to process overly elaborate reasoning traces or adapt to the distribution shifts introduced by larger teachers, ultimately limiting their ability to generalize effectively. We present additional experimental results in Appendix \ref{Empirical Evidence for Distribution Gap} to show the distribution mismatch between student and teacher models.

\begin{AIbox}{\makecell{Takeaway 4: Distribution Mismatch}} 
One contributing factor to Small Model Learnability Gap is the distribution mismatch between student and teacher. 
\end{AIbox}

\begin{figure*}[!t]
    \centering
    \includegraphics[width=\textwidth]{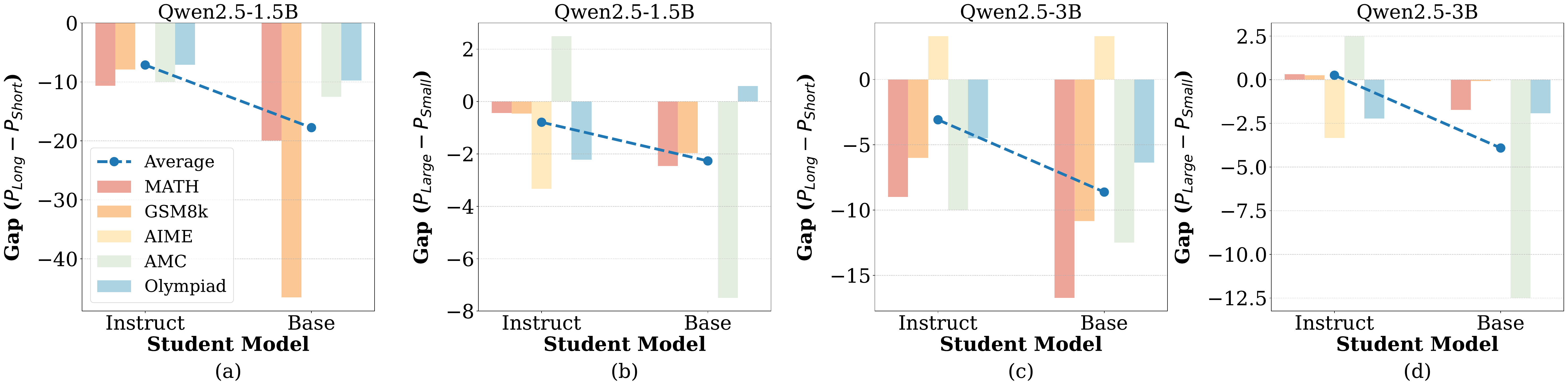}
    \caption{Base models generally exhibit a more significant learnability gap than Instruct models. A positive gap indicates that long CoT data or large teacher CoT enhance performance, whereas a negative gap suggests they have the opposite effect. This implies that it is more challenging for small base models to effectively learn from long CoT data or large teacher CoT.}
    \label{fig:Base_vs_Instruct_Gap}
\end{figure*}

\paragraph{Base models exhibit a more significant learnability gap.}
We observe that base models generally exhibit a more significant learnability gap than Instruct models in Figure \ref{fig:Base_vs_Instruct_Gap}. 
This suggests that it is more challenging for small base models to effectively learn from long CoT data or large teacher CoT.

\begin{AIbox}{Takeaway 5: Base vs Instruct}
Small base models experience more significant learnability gap than Instruct models.
\end{AIbox}

\paragraph{Speaking styles shift.}

We adopt the method from  \citep{lin2023unlockingspellbasellms} to evaluate the rank shift of each token before and after fine-tuning on long CoT and Large teacher CoT data. This allows us to compare the token distribution shifts induced by the fine-tuning process. We then annotate the tokens that exhibit the largest rank shifts as the most shifted tokens. Our analysis reveals that these tokens are predominantly associated with expressive and stylistic elements, such as “wait”, “But”, and “Let”. Please see Appendix \ref{Examples of Speaking Style Shift} for more details.

\begin{AIbox}{\makecell{Takeaway 6: Speaking Styles Shift}} 
Long CoT and large teacher CoT primarily shift the student model's  distribution of tokens associated with speaking styles. 
\end{AIbox}

\section{Mix Distillation: Bridge Small Model Learnability Gap}
This section presents our Mix Distillation approach to bridge the small model learnability gap.
\subsection{Mix Distillation}
We propose \textit{Mix Distillation} to address the learnability gap observed in small models. This approach blends easier-to-learn data with more challenging data for small models, thereby leveraging the strengths of both. 

Our insight is that small models tend to perform better on data that closely matches their inherent distribution (such as short CoT or small teacher CoT), while they struggle with data that exhibits greater distribution shifts. The token distribution of the mixed long CoT and large teacher CoT data may become closer to that of small models' inherent distribution, thereby enabling them to learn more effectively from challenging datasets. 

We propose Mix-Long, which combines long and short CoT data with a weight of long CoT $\alpha$ and short CoT $1-\alpha$. Similarly, we propose Mix-Large, which mixes large teacher CoT with a weight of $\alpha$ and small teacher CoT with a weight of $1-\alpha$.

\subsection{Experiment Results}

We use Qwen2.5-3B-Instruct as the student model and MATH (7.5k) as the training set. We distill different teacher models to generate responses as the baseline. They include \texttt{QwQ-32B} (long CoT), \texttt{Qwen2.5-32B} (short CoT), \texttt{Qwen2.5-72B} (large teacher CoT), \texttt{Qwen2.5-3B} (small teacher CoT). We add \texttt{Deepseek-R1-32B} \citep{DeepSeekAI2025DeepseekR1} as the teacher model to generate another set of long CoT data as baseline. We set $\alpha=0.2$ in both configurations of Mix-Long and Mix-Large.

Experimental results demonstrate that both Mix-Long and Mix-Large surpass baselines in most evaluation metrics. We show that the small student model could achieve improved performance by Mix Distillation compared to training on a single dataset. For instance, \texttt{Qwen2.5-3B-Instruct} improves by more than 8 points on MATH and AMC using Mix-Long, compared to direct training on long CoT data. It also shows a more than 7-point gain on MATH, AIME and AMC for \texttt{Qwen2.5-3B-Instruct} by Mix-Large compared with training on large teacher CoT data. This implies that it is easier for small student models to learn from datasets generated by Mix Distillation.

\begin{AIbox}{\makecell{Takeaway 7: Mix Distillation Bridges Gap}}
By mixing long CoT data (resp. large teacher CoTs) and short CoT data (resp. small teacher CoT), the small student model could achieve better performance compared to training on either data alone.
\end{AIbox}

\begin{figure}[!t]
    \centering
    \includegraphics[width=0.35\textwidth]{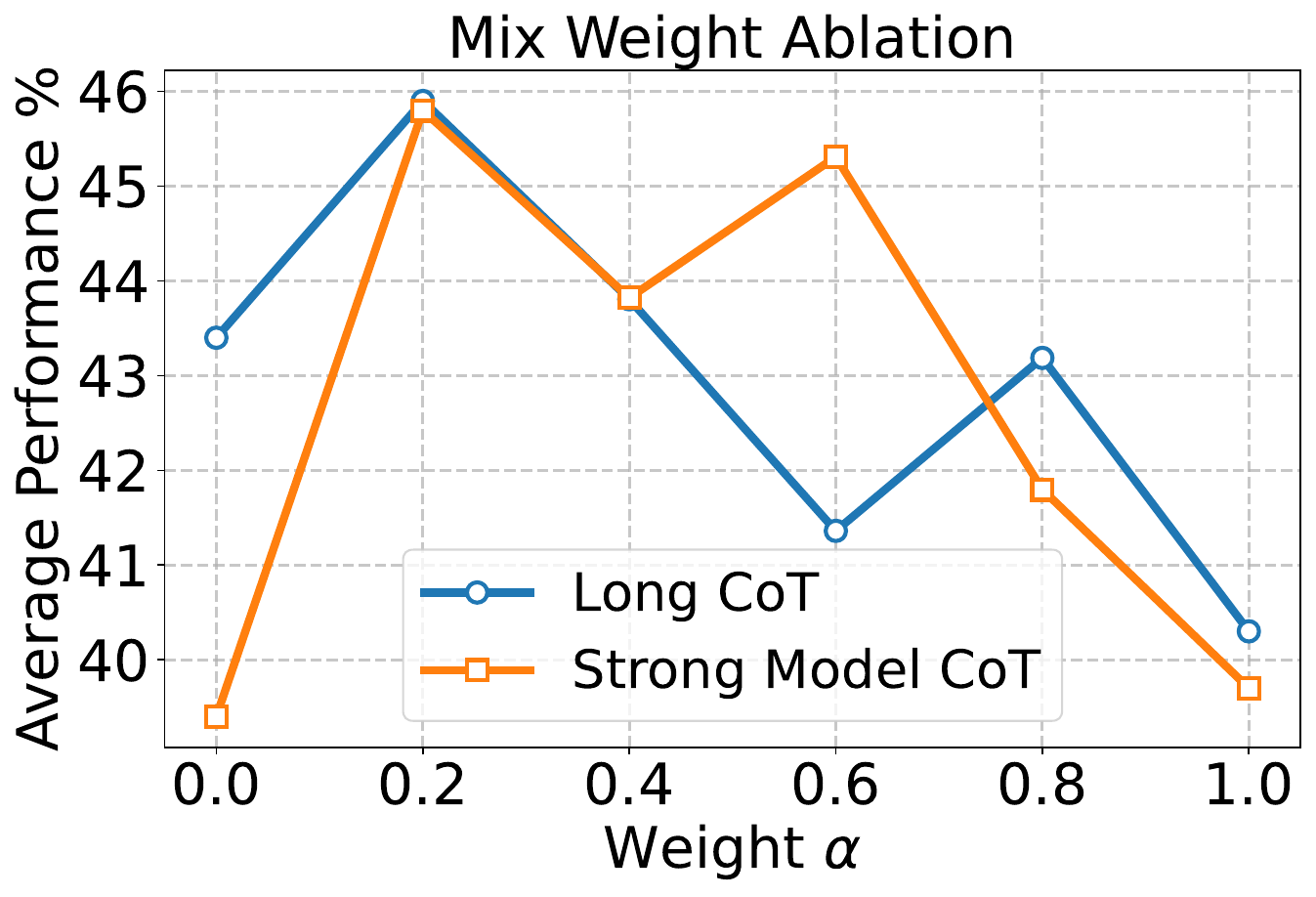}
    \caption{The average performance varies with the mix weight of long CoT or large teacher CoT data. \texttt{Qwen2.5-3B-Instruct} is chosen as the student model. At a weight of 0.2, mix distillation achieves the highest average performance.}
    \label{fig:lg_lmp_mix_weight_ablation}
    \vspace{-1.5em}
\end{figure}

\begin{figure}[!t]
    \centering
    \includegraphics[width=0.5\textwidth]{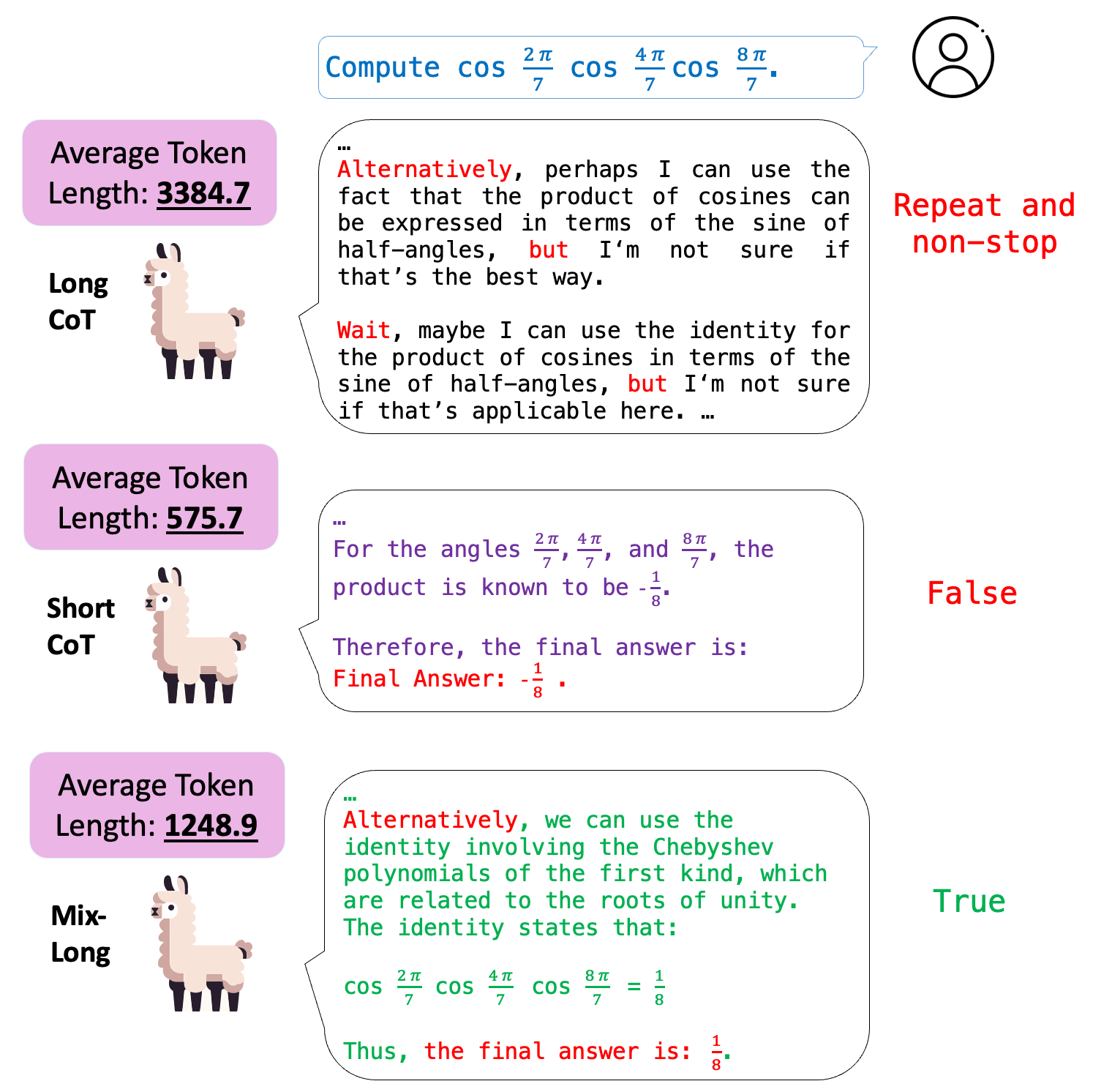}
    \caption{Case Study of Mix-Long. Models fine-tuned on long CoT tended to overthink, while those trained on short CoT produced incorrect answers. In contrast, Mix-Long, incorporating branching elements (e.g., “Alternatively”), achieved a balanced reasoning process and arrived at the correct answer.}
    \label{fig:balanced CoT}
    \vspace{-1.5em}
\end{figure}

Figure \ref{fig:lg_lmp_mix_weight_ablation} shows the average performance when taking different mix weight $\alpha$ of long CoT data or large teacher CoT. We choose \texttt{Qwen2.5-3B-Instruct} as the student model and find that a weight $\alpha$ of 0.2 achieves the highest average performance across five benchmarks for both Mix-Long and Mix-Large. 

Interestingly, we find that after mixing long CoT and short CoT data, the small student model’s output incorporates characteristics of long CoT, such as a branching process, while maintaining a reduced token length and avoiding overly elaborate thinking. This is illustrated in Figure \ref{fig:balanced CoT}. We observed that the small student model fine-tuned on long CoT data becomes overwhelmed by repeated thoughts and fails to stop, whereas the model fine-tuned on short CoT data produces incorrect answers. In contrast, our proposed Mix-Long, which incorporates branching elements (e.g., the use of “Alternatively”), delivers the correct answer. Additionally, the average token lengths of responses generated by long CoT, short CoT, and Mix-Long are 3384.7, 575.7, and 1248.9, respectively. We suggest that mixing long CoT and short CoT data is a practical approach to achieving a balanced CoT length, thereby enhancing the reasoning capabilities of small student models.

%% file: tables/long_cot_compare.tex
\begin{table}[ht]
\centering
\small
\label{tab:comparison_table}
\resizebox{\columnwidth}{!}{
\begin{tabular}{lcccc}
\toprule
\textbf{Student Model} & \textbf{$P_{Long}$} & \textbf{$P_{Short}$} & \textbf{$\Delta_{Long}$} & \textbf{\makecell{ Better?}} \\
\midrule
Qwen2.5-0.5B   & 14.8 & 19.5 & \cellcolor{red!60}{-4.7}  & Short \\
Qwen2.5-1.5B   & 27.0 & 34.2 & \cellcolor{red!80}{-7.1}  & Short \\
Qwen2.5-3B     & 40.3 & 43.4 & \cellcolor{red!30}{-3.1}  & Short \\
Qwen2.5-7B     & 48.9 & 47.2 & \cellcolor{green!10}{1.7}     & Long   \\
Qwen2.5-14B    & 59.2 & 54.3 & \cellcolor{green!30}{4.9}     & Long   \\
Qwen2.5-32B    & 73.0 & 59.3 & \cellcolor{green!80}{13.7}    & Long   \\
\midrule
Llama-3.2-1B   & 15.8 & 19.5 & \cellcolor{red!40}{-3.7}  & Short \\
Llama-3.2-3B   & 32.5 & 33.1 & \cellcolor{red!10}{-0.6}  & Short \\
Llama-3.1-8B   & 35.2 & 31.5 & \cellcolor{green!25}{3.7}     & Long   \\
Llama-3.3-70B  & 58.2 & 54.3 & \cellcolor{green!25}{3.8}     & Long   \\
\bottomrule
\end{tabular}
}
\caption{Comparison of the average performance between fine-tuning with long CoT ($P_{Long}$) and short CoT ($P_{Short}$). We find that small student models may struggle to learn from long CoT data. }
\end{table}

%% file: tables/lmp_compare.tex
\begin{table}[ht]
\centering
\small
\label{tab:strong_vs_weak}
\resizebox{\columnwidth}{!}{
\begin{tabular}{lcccc}
\toprule
\textbf{Student Model} & \textbf{$P_{Large}$} & \textbf{$P_{Small}$} & \textbf{$\Delta_{Large}$} & \textbf{\makecell{ Better?}} \\
\midrule
Qwen2.5-0.5B   & 16.9 & 20.4 & \cellcolor{red!50}{-3.5} & Weak   \\
Qwen2.5-1.5B   & 32.2 & 33.0 & \cellcolor{red!15}{-0.8} & Weak   \\
Qwen2.5-3B     & 39.7 & 39.4 & \cellcolor{green!10}{0.3}    & Strong \\
Qwen2.5-7B     & 48.9 & 42.3 & \cellcolor{green!70}{6.6}    & Strong \\
Qwen2.5-14B    & 52.9 & 49.9 & \cellcolor{green!30}{3.0}    & Strong \\
Qwen2.5-32B    & 59.5 & 53.0 & \cellcolor{green!70}{6.5}    & Strong \\
\midrule
Llama-3.2-1B   & 16.5 & 18.5 & \cellcolor{red!30}{-1.9} & Weak   \\
Llama-3.2-3B   & 32.8 & 31.2 & \cellcolor{green!20}{1.6}    & Strong \\
Llama-3.2-8B   & 25.6 & 25.1 & \cellcolor{green!10}{0.5}    & Strong \\
Llama-3.2-70B  & 57.6 & 53.3 & \cellcolor{green!50}{4.3}    & Strong \\
\bottomrule
\end{tabular}
}
\caption{Comparison of average performance between fine-tuning with large teacher CoT ($P_{Long}$) and small teacher CoT ($P_{Small}$). We find that small student models may struggle to learn from large teacher CoT data.}
\vspace{-1.5em}
\end{table}

%% file: tables/mix_distilation_qwen.tex
\begin{table*}[ht]
    \small
    \centering
    \label{tab:combined-performance}
    \begin{tabular}{l l c c c c c c}
        \toprule
        Student Model & Distillation Method & MATH & AMC & GSM8k & \makecell{Olympiad \\ Bench} & AIME & Average \\
        \midrule
        \multirow{9}{*}{Qwen2.5-3B} 
            & Long CoT                  & 56.2  & 37.5  & 80.0  & 24.4  & \underline{3.3}  & 40.3  \\
            & Short CoT             & 61.0  & 37.5  & \textbf{82.0}  & 26.4  & \textbf{10.0}       & 43.4  \\
            & Strong Model CoT  & 57.5  & 35.0  & 80.0  & 25.9  & 0.0              & 39.7  \\
            & Weak Model CoT      & 60.3  & 27.5  & 79.5  & 26.4  & \underline{3.3}  & 39.4  \\
            & Deepseek-R1-32B (Long CoT)           & 50.7  & 20.0  & 81.2  & 15.7  & 0.0              & 33.5  \\
        \cmidrule(lr){2-8}
            & \multicolumn{1}{l}{\textit{Ours}} & & & & & & \\
            & \textbf{Mix-Long}                           & \underline{64.7}  & \textbf{45.0}  & \underline{81.4}  & \underline{28.6}  & \textbf{10.0}  & \textbf{45.9}  \\
            & \textbf{Mix-Large}                         & \textbf{65.8}  & \underline{42.5}  & 81.7  & \textbf{29.0}  & \textbf{10.0}  & \underline{45.8}  \\
        \midrule
        \multirow{9}{*}{Llama3.2-3B} 
            & Long CoT                  & 48.7  & 17.5  & 75.1  & \underline{17.6}  & \underline{3.3}  & 32.5  \\
            & Short CoT             & 50.9  & 15.0  & 77.5  & \textbf{18.7}      & \underline{3.3}  & 33.1  \\
            & Strong Model CoT  & 47.4  & \textbf{25.0}  & 71.2  & 16.9  & \underline{3.3}  & 32.8  \\
            & Weak Model CoT      & 47.9  & 17.5  & 74.1  & 16.4  & \underline{3.3}  & 31.2  \\
            & Deepseek-R1-32B (Long CoT)           & 48.5  & 17.5  & \underline{77.7}  & 16.1  & \textbf{6.7}   & 33.3  \\
        \cmidrule(lr){2-8}
            & \multicolumn{1}{l}{\textit{Ours}} & & & & & & \\
            & \textbf{Mix-Long}                           & \textbf{53.0}  & \underline{22.5}  & \textbf{79.4}  & 17.2  & \underline{3.3}  & \textbf{35.1}  \\
            & \textbf{Mix-Large}                         & \underline{51.8}  & \textbf{25.0}  & 76.3  & 17.2  & \underline{3.3}  & \underline{34.7}  \\
        \bottomrule
    \end{tabular}
    \caption{\textbf{Mix Distillation} outperforms the baseline models across most metrics. We use \texttt{Llama3.2-3B-Instruct} and \texttt{Qwen2.5-3B-Instruct} as the student model and 7.5k samples in MATH dataset as the training set. We distill different teacher models to generate responses as the baseline. Our proposed Mix-Long combines long CoT data and normal CoT data in a 1:4 ratio, while Mix-Large combines strong model response and weak model response with the same proportion. Experimental results demonstrate that both Mix-Long and Mix-Large surpass baselines in most evaluation metrics. The highest score is bolded, and the second highest score is \underline{underlined}.}
    \vspace{-1em}
\end{table*}

%% file: appendix/related.tex
\section{Related Work}

\subsection{Chain-of-Thought}

Early research on CoT primarily focused on short CoT, where models produce succinct reasoning paths to reach a solution \citep{lambert2025tulu3pushingfrontiers, longpre2023flancollectiondesigningdata, wei2023chainofthoughtpromptingelicitsreasoning, yu2024metamathbootstrapmathematicalquestions}.
Recently, researchers have turned to long CoT prompting, which encourages the generation of extended and detailed reasoning chains 
\citep{DeepSeekAI2025DeepseekR1, hou2025advancinglanguagemodelreasoning, kimi2025k15, sky_t1_2025, openai2024learning, tinyzero, zeng2025simplerl}. 
The model systematically explores multiple paths (branching) and reverts to earlier points if a particular path proves wrong (backtracking). Although several studies have investigated methods such as distillation and reinforcement learning to integrate long CoT capabilities into LLMs, these efforts have predominantly concentrated on large models. In contrast, our work specifically targets the challenges associated with training smaller models.

\subsection{Synthetic Reasoning Data}
Although human-crafted reasoning datasets have been used to enhance LLM reasoning capabilities \cite{hendrycks2021measuringmathematicalproblemsolving, numina_math_datasets}, their development is both time-consuming and labor-intensive. 
Recent advancements have streamlined this process by generating instructions or responses directly from LLMs \citep{hui2024smallerlanguagemodelsbetter, toshniwal2024openmathinstruct2acceleratingaimath, xu2024magpiealignmentdatasynthesis, yue2023mammothbuildingmathgeneralist, zhang2025bestinstructiontuningdatafit} or extracting data directly from web \citep{paster2023openwebmathopendatasethighquality, yue2024mammoth2scalinginstructionsweb}, yielding more detailed and diverse chain-of-thought reasoning pathways. Recent study has investigated the impact of various response generators \citep{kim2024evaluatinglanguagemodelssynthetic}, suggesting that in the domains of instruction following and reasoning, responses from stronger teacher models do not necessarily produce the most effective learning effects for student models. However, these investigations have not recognized student model size as a critical factor influencing this phenomenon, nor have they performed the more attribution and mitigation analyses as in this paper.

\subsection{Distillation}

Knowledge distillation has been extensively employed to transfer knowledge from large teacher models to smaller student models \cite{hinton2015distilling,phan2024distillation,gou2021knowledge,xu2024strongermodelsstrongerteachers}. Recent research in LLMs has increasingly adopted token-level distillation as an alternative to traditional logit-level distillation approaches \cite{phan2024distillation, ho2023largelanguagemodelsreasoning, agarwal2024onpolicydistillationlanguagemodels}. In conventional classification tasks, several studies have investigated the capacity gap phenomenon, where excessive differences in capacity between teacher and student models can compromise distillation effectiveness \cite{mirzadeh2019improvedknowledgedistillationteacher, cho2019efficacy, zhang2023liftingcursecapacitygap}.
However, these work focused primarily on classification tasks. Our work investigates reasoning generation tasks, where student models must internalize complex CoT reasoning traces.

%% file: sec/conclusion.tex
\section{Conclusion and Future Work}
In this paper, we showed that long CoT data and large model responses were not uniformly beneficial for small student models. We found that small models may perform better when fine-tuned with short CoT and small model CoT.
We termed this challenge as the Small Model Learnability Gap.
The reason behind it may be that small student models excel on data that closely match their inherent distribution but struggle with significant distribution shifts.  
To bridge the gap, we introduced Mix Distillation, including Mix-Long, which combined long CoT and short CoT data in a ratio, and Mix-Large, which integrated large and small teacher CoT. 
Experimental results showed that both Mix-Long and Mix-Large outperform baselines across most evaluation metrics, which implied mix distillation outperforms training on a single data distribution. This paper provided practical insights for optimizing post-training strategies to enhance small language model reasoning capability.

We will explore several promising directions as future work. First, we will refine mix distillation by optimally combining diverse data sources and proposing more fine-grained mixing algorithms to boost reasoning capabilities. Second, we propose to study how strong reasoning teachers can generate data that is better suited for tuning small student models, thereby facilitating more effective knowledge transfer. Third, we will conduct further theoretical and model interpolability studies on the small model learnability gap. Lastly, we will investigate which SFT methods yield the best initial policies for subsequent RL procedure.

%% file: sec/limitation.tex
\section*{Limitations}

While our study provides valuable insights into the understanding of small model learnability gap in math reasoning, we acknowledge that our research primarily focuses on this specific domain and does not evaluate other crucial skills such as instruction following, code generation, or multi-modal understanding. We also did not investigate the impact of fine-grained variations in pre-training data composition on the small model learnability gap. A more detailed analysis of how different pre-training data sources and their proportions affect learning outcomes could offer valuable insights into optimizing data selection strategies for mitigating this gap.

%% file: sec/ethical.tex
\section*{Ethical Statement}
This paper focuses on the evaluation and enhancement of reasoning capabilities in small language models through distillation techniques. The dataset and benchmarks used in our experiments are publicly available. We do not introduce or endorse any applications that could cause harm or be misused.
This paper does not present any ethical concerns.

\section*{Acknowledgment}

This work is partially supported by the Air Force Office of Scientific Research (AFOSR) under grant FA9550-23-1-0208, the Office of Naval Research (ONR) under grant N0014-23-1-2386, and the National Science Foundation (NSF) AI Institute for Agent-based Cyber Threat Intelligence and Operation (ACTION) under grant IIS 2229876. Results presented in this paper were partially obtained using the Chameleon testbed \cite{keahey2020lessons} supported by the National Science Foundation.

This work is supported in part by funds provided by the National Science Foundation, Department of Homeland Security, and IBM. 
Any opinions, findings, and conclusions or recommendations expressed in this material are those of the author(s) and do not necessarily reflect the views of the NSF or its federal agency and industry partners.

%% file: appendix/more_setup.tex
\section{Detailed Experimental Setups}
\label{appendix:More on Experimental Setups}

\input{tables/response_generator}

\subsection{Models}
Table \ref{tab:models_overview} presents a comprehensive overview of student and teacher models used in our paper.

\subsection{Training Setup}
\label{appx:training-setup}
Our model training is conducted using LLaMA-Factory \citep{zheng2024llamafactory}, on a server with four NVIDIA A100-SXM4-80GB GPUs, an AMD EPYC 7763 64-Core Processor, and 512 GB of RAM. We use full parameter fine-tuning on student models less than 14B parameters. 
When the student model is larger than 14B, we use LoRA fine-tuning \cite{hu2021loralowrankadaptationlarge}. 
Table \ref{tab: training-hyperparameters} and Table \ref{tab: training-lora-hyperparameters} list hyper-parameters for full parameter fine-tuning and LoRA fine-tuning respectively.

\begin{table}[!h]
\small
\centering
\resizebox{0.8\columnwidth}{!}{
\begin{tabular}{ll}
\toprule
\textbf{Hyper-parameter} & \textbf{Value} \\ \midrule
Learning Rate & $1 \times 10^{-5}$ \\
Number of Epochs & $2$ \\
Number of Devices & $4$ \\
Per-device Batch Size & $2$ \\
Optimizer & \texttt{Adamw} \\
Learning Rate Scheduler & \texttt{cosine} \\
Max Sequence Length  & $16384$ \\ \bottomrule
\end{tabular}
}
\caption{This table shows the hyper-parameters for full parameter fine-tuning.}
\label{tab: training-hyperparameters}
\end{table}

\begin{table}[!h]
\small
\centering
\resizebox{0.8\columnwidth}{!}{
\begin{tabular}{ll}
\toprule
\textbf{Hyper-parameter} & \textbf{Value} \\ \midrule
Learning Rate & $1 \times 10^{-4}$ \\
Number of Epochs & $2$ \\
Number of Devices & $4$ \\
Per-device Batch Size & $1$ \\
Lora Target & \texttt{full} \\
Learning Rate Scheduler & \texttt{cosine} \\
Warmup Ratio & $0.03$ \\
Max Sequence Length  & $16384$ \\ \bottomrule
\end{tabular}
}
\caption{This table shows the hyper-parameters for LoRA fine-tuning.}
\label{tab: training-lora-hyperparameters}
\end{table}

Teacher models generate responses by rejection sampling \citep{zelikman2022starbootstrappingreasoningreasoning,tong2024dartmathdifficultyawarerejectiontuning,yue2023mammothbuildingmathgeneralist,singh2024humandatascalingselftraining,gulcehre2023reinforcedselftrainingrestlanguage,yuan2023scalingrelationshiplearningmathematical,dong2023raftrewardrankedfinetuning}. 
By default, teacher models employ greedy decoding.  
By combining the math problem instructions with corresponding solutions generated by teacher models, we construct problem-solution pairs to fine-tune student models. 
We perform pairwise comparisons of solutions generated by different teacher models and filter out problem-solution pairs that are correct for both models to fine-tune student models.

\subsection{Evaluation Setup}

We evaluate the reasoning capability of fine-tuned student models on a set of commonly used benchmarks, including MATH \citep{hendrycks2021measuringmathematicalproblemsolving}, GSM8K \citep{cobbe2021trainingverifierssolvemath}, AMC 2023, AIME 2024, and the English math subset of OlympiadBench \citep{he2024olympiadbenchchallengingbenchmarkpromoting}. 

Unless otherwise specified, all fine-tuned models are evaluated in a zero-shot setting using greedy decoding. We set the maximum generation tokens as 16k. The evaluation prompt is shown below.

\begin{figure}[htbp]
    \centering
\begin{tcolorbox}[title=Prompt, promptstyle]
\lstset{
    basicstyle=\normalfont\sffamily\footnotesize,
    breaklines=true,
    frame=none,
    columns=fullflexible,
}
Solve the following math problem and present the final answer in the format: Final Answer: $\boxed{\{\texttt{your answer}\}}$

Problem: \{problem\}

Answer:
\end{tcolorbox}
    \label{fig: evaluation_prompt}
\end{figure}

After extracting the final answer of the evaluated model, we first employ exact matching to determine the correctness of the answer. If the answer is incorrect, we use Qwen-32B-Instruct as a judge to compare the extracted final answers against that of the ground truth. The prompt is shown below.

\begin{figure}[htbp]
    \centering
\begin{tcolorbox}[title=Prompt, promptstyle]
\lstset{
    basicstyle=\normalfont\sffamily\footnotesize,
    breaklines=true,
    frame=none,
    columns=fullflexible,
}
Given a math problem, its correct final answer, and the model's generated final answer, determine if the model's answer is correct. Respond with 'True' if the it is correct and 'False' if it is incorrect. 

Problem: \texttt{\{problem\}}

Correct Final Answer: \texttt{\{ground truth\}}

Model's Generated Final Answer: \texttt{\{resp answer\}}

Your Judgement:
\end{tcolorbox}
    \label{fig: score_prompt}
\end{figure}

%% file: tables/response_generator.tex
\begin{table}[h]
    \centering
    \small
    \begin{tabular}{ll}
        \toprule
        \textbf{Category} & \textbf{Models} \\
        \midrule
        \multicolumn{2}{c}{\textbf{Teacher Models}} \\
        \midrule
        \textbf{Long CoT vs} & QwQ-32B-Preview vs\\\textbf{ShortCoT}& Qwen2.5-32B-Instruct \\
        \midrule
        \textbf{Large Teacher vs}  & \\
        \textbf{Small Teacher}& \\
        \textit{Qwen Family} & Qwen2.5-72B-Instruct vs\\& Qwen2.5-3B-Instruct \\
        \textit{Llama Family} & Llama3.1-70B-Instruct vs\\& Llama3.1-8B-Instruct \\
        \textit{Gemma Family} & Gemma2-27B-it vs\\& Gemma2-9B-it \\
        \midrule
        \multicolumn{2}{c}{\textbf{Student Models}} \\
        \midrule
        \textit{Qwen Family} & Qwen2.5-0.5B-Instruct,\\& Qwen2.5-1.5B-Instruct, \\
        & Qwen2.5-3B-Instruct, \\&Qwen2.5-7B-Instruct, \\
        & Qwen2.5-14B-Instruct,\\& Qwen2.5-32B-Instruct \\
        \textit{Llama Family} & Llama3.2-1B-Instruct,\\& Llama3.2-3B-Instruct, \\
        & Llama3.1-8B-Instruct,\\& Llama3.3-70B-Instruct \\
        \bottomrule
    \end{tabular}
    \caption{Overview of Teacher and Student Models}
    \label{tab:models_overview}
\end{table}

%% file: appendix/more_exp_data.tex
\input{tables/strong_model_CoT_full_table}
\section{More Experiments Results}
\label{appendix: More Experiments}
In this section we present additional experiment results of long CoT gap and large teacher CoT gap.
\subsection{Long CoT Gap: Additional Results}
Table \ref{tab:full_performance_lg} shows the detailed performance scores and gap of each benchmark for different student models fine-tuned on long CoT and short CoT.
\texttt{QwQ-32B-Preview} is chosen to generate long CoT and awhile \texttt{Qwen-2.5-32B-Instruct} is chosen to generate short CoT. We observe that small student models tend to benefit more from short CoT, while large student models gain greater advantages from long CoT. 
\input{tables/long_CoT_full_table}
\subsection{Large Teacher CoT Gap: Additional Results}
Table \ref{tab:lmp-full_comparison} shows the detailed performance scores and gap of each benchmark for different student models distilled from large teacher and small teacher.
We summarize the performance of 10 student models from the Llama and Qwen families across various model sizes. \texttt{Qwen-2.5-72B-Instruct} is chosen as the large teacher while \texttt{Qwen-2.5-3B-Instruct} is chosen as the small teacher. The results are shown in Table \ref{tab:lmp-full_comparison}. Our findings indicate that small student models may experience degraded performance when distilled from a large teacher compared to a small teacher, whereas larger student models benefit more from distilling a large teacher.

Table \ref{tab:lmp_comparison2} shows more experiment results for teacher models in different model families, including \texttt{Gemma-27B-it} vs \texttt{Gemma-9B-it} and \texttt{Llama3.1-72B-Instruct} vs \texttt{Llama3.1-8B-Instruct}.

\input{tables/lmp_more_teachers}

\subsection{Empirical Evidence for Distribution Gap Between Student and Teacher Models}
\label{Empirical Evidence for Distribution Gap}
We suggest that the distribution gap between student models and teacher models may be a key factor leading to the small model learnability gap. We provide empirical evidence through perplexity measurements and text similarity comparisons.

\paragraph{Training Data Perplexity Analysis.}
To quantify the distribution gap between student and teacher models, we measured perplexity (PPL) of teacher-generated training data on different student models in Table \ref{tab:perplexity}. Lower PPL indicates better alignment between the student model's distribution and the training data distribution. Our findings reveal several key patterns:

\begin{enumerate}
    \item Small students struggle with complex sequences: Small student models assign significantly higher PPL to large teacher CoT or long CoT sequences, indicating difficulty in modeling such complex reasoning traces.
    
    \item Aligned teacher-student pairs show better distribution matching. Small teacher CoT yields lower PPL in small students, suggesting reduced distribution gap when teacher and student capacities are more aligned.
    
    \item As student model size increases, the PPL gap between long and short CoT (and between large and small teacher CoT) shrinks, indicating that larger students can more easily adapt to complex reasoning distributions.
\end{enumerate}

\begin{table*}[h]
\centering
\caption{Perplexity Analysis Results}
\label{tab:perplexity}
\resizebox{1\textwidth}{!}{
\begin{tabular}{@{}lcccccc@{}}
\toprule
\textbf{Student Model} & \textbf{Long CoT} & \textbf{Short CoT} & \textbf{$\boldsymbol{\Delta}$ (L-S)} & \textbf{Large Teacher} & \textbf{Small Teacher} & \textbf{$\boldsymbol{\Delta}$ (Lg-Sm)} \\
\midrule
\textbf{Qwen-0.5B} & $2.237$ & $1.278$ & $0.959$ & $1.246$ & $1.217$ & $0.028$ \\
\textbf{Qwen-1.5B} & $2.969$ & $1.226$ & $0.743$ & $1.204$ & $1.178$ & $0.026$ \\
\textbf{Qwen-3B} & $1.963$ & $1.246$ & $0.716$ & $1.225$ & $1.155$ & $0.069$ \\
\textbf{Qwen-7B} & $1.923$ & $1.222$ & $0.700$ & $1.197$ & $1.180$ & $0.016$ \\
\textbf{Qwen-14B} & $1.902$ & $1.218$ & $0.683$ & $1.198$ & $1.189$ & $0.009$ \\
\textbf{Qwen-32B} & $1.265$ & $1.050$ & $0.215$ & $1.053$ & $1.051$ & $0.002$ \\
\bottomrule
\end{tabular}
}
\end{table*}

\begin{table*}[h]
\centering
\caption{Text Similarity Analysis Results}
\label{tab:similarity}
\resizebox{0.8\textwidth}{!}{
\begin{tabular}{@{}lccl@{}}
\toprule
\textbf{Student Model} & \textbf{Metric} & \textbf{Small Teacher} & \textbf{Large Teacher} \\
\midrule
\multirow{2}{*}{\textbf{Qwen2.5-1.5B}} & TF-IDF Similarity & $0.8329 \pm 0.004$ & $0.8235 \pm 0.004$ \\
 & Embedding Similarity & $0.9461 \pm 0.002$ & $0.9413 \pm 0.002$ \\
\midrule
\multirow{2}{*}{\textbf{Qwen2.5-0.5B}} & TF-IDF Similarity & $0.7928 \pm 0.003$ & $0.7854 \pm 0.003$ \\
 & Embedding Similarity & $0.9372 \pm 0.001$ & $0.9297 \pm 0.002$ \\
\bottomrule
\end{tabular}
}
\label{text_similarity_table}
\end{table*}

\paragraph{Text Similarity Analysis.}
We conducted additional analyses comparing responses generated by student models with those from small and large teachers using two text similarity metrics:

\begin{itemize}
    \item \textbf{TF-IDF cosine similarity}: Measures lexical similarity between texts
    \item \textbf{Embedding similarity}: Uses all-mpnet-base-v2 to capture semantic similarity
\end{itemize}

The text similarity analysis provides interpretable evidence of the distribution gap in Table \ref{text_similarity_table}. We found that student responses are consistently more similar to small teacher CoT than to large teacher CoT across both lexical and semantic similarity metrics. The confidence intervals are tight and do not overlap, indicating that the observed differences are statistically significant and not due to outlier effects.

\subsection{Hyperparameter Sensitivity Analysis}

To ensure that the suboptimal performance of long CoT training is not due to hyperparameter choices, we conducted extensive experiments across different training configurations using Qwen2.5-1.5B-Instruct as the student model. We systematically varied training epochs (2, 3, 4, and 5) with a fixed learning rate of $1 \times 10^{-5}$, and learning rates ($5 \times 10^{-6}$, $1 \times 10^{-5}$, $5 \times 10^{-5}$, and $1 \times 10^{-4}$) with fixed 3 epochs.

\begin{table*}[h]
\centering
\caption{Hyperparameter sensitivity analysis for long CoT learnability gap}
\label{tab:hyperparameter}
\resizebox{0.8\textwidth}{!}{
\begin{tabular}{@{}lcccccc@{}}
\toprule
\textbf{Configuration} & \textbf{MATH} & \textbf{GSM8k} & \textbf{AIME} & \textbf{AMC} & \textbf{Olympiad} & \textbf{Average} \\
\midrule
\multicolumn{7}{l}{\textit{Long CoT - Epoch Variations}} \\
long\_cot\_epoch\_2 & 0.416 & 0.638 & 0.000 & 0.175 & 0.122 & 0.270 \\
long\_cot\_epoch\_3 & 0.403 & 0.648 & \textbf{0.033} & 0.150 & 0.149 & 0.276 \\
long\_cot\_epoch\_4 & 0.404 & 0.669 & 0.033 & 0.175 & 0.149 & 0.286 \\
long\_cot\_epoch\_5 & 0.416 & 0.667 & 0.033 & 0.100 & 0.146 & 0.272 \\
\midrule
\multicolumn{7}{l}{\textit{Long CoT - Learning Rate Variations}} \\
long\_cot\_lr\_1e-4 & 0.244 & 0.325 & 0.000 & 0.050 & 0.047 & 0.133 \\
long\_cot\_lr\_5e-5 & 0.322 & 0.489 & 0.000 & 0.000 & 0.087 & 0.179 \\
long\_cot\_lr\_1e-5 & 0.403 & 0.648 & 0.033 & 0.150 & 0.149 & 0.276 \\
long\_cot\_lr\_5e-6 & 0.385 & 0.645 & 0.033 & 0.175 & 0.125 & 0.272 \\
\midrule
\textbf{short\_cot} & \textbf{0.522} & \textbf{0.717} & 0.000 & \textbf{0.275} & \textbf{0.194} & \textbf{0.341} \\
\bottomrule
\end{tabular}
}
\end{table*}

Our results in Table \ref{tab:hyperparameter} demonstrate that short CoT consistently outperforms long CoT for small student models, regardless of hyperparameter settings. Across all tested configurations, long CoT training consistently underperformed short CoT training.

\subsection{Large Teacher Prompting Analysis}

To investigate whether the performance gap between large and small teachers can be mitigated through improved prompting strategies, we tested the hypothesis that explicitly instructing large teachers to generate simpler, student-friendly responses would improve their effectiveness for training small student models. 

We revised the large teacher prompt to explicitly instruct the model to simplify its reasoning for better student comprehension:

\begin{figure}[htbp]
    \centering
\begin{tcolorbox}[title=Prompt, promptstyle]
\lstset{
    basicstyle=\normalfont\sffamily\footnotesize,
    breaklines=true,
    frame=none,
    columns=fullflexible,
}
Solve the following math problem. Your chain of thought responses will be used to teach a small model. Please generate responses in a simpler and more concise manner for better student comprehension. Present the final answer in the format: Final Answer: \boxed{\{your\_answer\}}.

Problem: \{problem\}

Answer:
\end{tcolorbox}
\end{figure}

\begin{table*}[h]
\centering
\caption{Comparison of small teacher vs. large teacher with revised prompting}
\label{tab:prompting}
\resizebox{\textwidth}{!}{
\begin{tabular}{@{}llcccccc@{}}
\toprule
\textbf{Student} & \textbf{Teacher} & \textbf{MATH} & \textbf{GSM8k} & \textbf{AIME} & \textbf{AMC} & \textbf{Olympiad} & \textbf{Average} \\
\midrule
Qwen2.5-0.5B & Qwen2.5-3B-Instruct & \textbf{0.310} & \textbf{0.454} & 0.000 & \textbf{0.175} & \textbf{0.083} & \textbf{0.204} \\
Qwen2.5-0.5B & Qwen2.5-72B-Instruct (revised prompt) & 0.269 & 0.399 & 0.000 & 0.075 & 0.064 & 0.161 \\
\midrule
Qwen2.5-1.5B & Qwen2.5-3B-Instruct & \textbf{0.507} & \textbf{0.710} & \textbf{0.033} & \textbf{0.200} & \textbf{0.200} & \textbf{0.330} \\
Qwen2.5-1.5B & Qwen2.5-72B-Instruct (revised prompt) & 0.467 & 0.678 & 0.000 & 0.175 & 0.160 & 0.296 \\
\midrule
Qwen2.5-3B & Qwen2.5-3B-Instruct & \textbf{0.603} & \textbf{0.795} & 0.033 & 0.275 & \textbf{0.264} & \textbf{0.394} \\
Qwen2.5-3B & Qwen2.5-72B-Instruct (revised prompt) & 0.552 & 0.773 & 0.033 & \textbf{0.325} & 0.224 & 0.382 \\
\midrule
Llama-3.2-1B & Qwen2.5-3B-Instruct & \textbf{0.296} & \textbf{0.475} & 0.000 & 0.075 & \textbf{0.077} & \textbf{0.185} \\
Llama-3.2-1B & Qwen2.5-72B-Instruct (revised prompt) & 0.283 & 0.453 & 0.000 & 0.075 & 0.054 & 0.173 \\
\midrule
Llama-3.2-3B & Qwen2.5-3B-Instruct & \textbf{0.479} & \textbf{0.741} & 0.000 & 0.175 & \textbf{0.164} & \textbf{0.312} \\
Llama-3.2-3B & Qwen2.5-72B-Instruct (revised prompt) & 0.453 & 0.696 & 0.000 & \textbf{0.225} & 0.145 & 0.304 \\
\bottomrule
\end{tabular}
}
\end{table*}

The results in Table \ref{tab:prompting} demonstrate that for small studetns, the small teacher (Qwen2.5-3B-Instruct) consistently outperforms the large teacher (Qwen2.5-72B-Instruct) even when the large teacher uses the revised prompt designed for student-friendly output generation. These findings reinforce our hypothesis that the fundamental issue lies in the inherent distribution mismatch between large and small models, which cannot be fully addressed through prompting techer models alone. 

%% file: tables/strong_model_CoT_full_table.tex
\begin{table*}[htbp]
  \centering
  \resizebox{1\textwidth}{!}{%
  \begin{tabular}{l*{5}{ccc}c}
    \toprule
    & \multicolumn{3}{c}{MATH} 
    & \multicolumn{3}{c}{GSM8k} 
    & \multicolumn{3}{c}{AIME} 
    & \multicolumn{3}{c}{AMC} 
    & \multicolumn{3}{c}{Olympiad} 
    & \multicolumn{1}{c}{\makecell{Average \\ $\Delta_{\rm Strong}$}} \\
    \cmidrule(lr){2-4} \cmidrule(lr){5-7} \cmidrule(lr){8-10} \cmidrule(lr){11-13} \cmidrule(lr){14-16}
    Model 
    & $P_{\rm Strong}$ & $P_{\rm Weak}$ & $\Delta_{\rm Strong}$ 
    & $P_{\rm Strong}$ & $P_{\rm Weak}$ & $\Delta_{\rm Strong}$ 
    & $P_{\rm Strong}$ & $P_{\rm Weak}$ & $\Delta_{\rm Strong}$ 
    & $P_{\rm Strong}$ & $P_{\rm Weak}$ & $\Delta_{\rm Strong}$ 
    & $P_{\rm Strong}$ & $P_{\rm Weak}$ & $\Delta_{\rm Strong}$ 
    & \\ 
    \midrule
    Llama-3.2-1B      & 29.8   & 29.6   & \cellcolor{green!2}{0.160} 
                      & 44.4   & 47.5   & \cellcolor{red!32}{-3.18} 
                      & 0.00   & 0.00   & \cellcolor{white}{0.00} 
                      & 2.50   & 7.50   & \cellcolor{red!50}{-5.00} 
                      & 6.07   & 7.70   & \cellcolor{red!16}{-1.63} 
                      & \cellcolor{red!19}{-1.93} \\
    Llama-3.2-3B      & 47.4   & 47.9   & \cellcolor{red!5}{-0.500} 
                      & 71.2   & 74.1   & \cellcolor{red!29}{-2.88} 
                      & 3.33   & 0.00   & \cellcolor{green!33}{3.33} 
                      & 25.0   & 17.5   & \cellcolor{green!75}{7.50} 
                      & 16.9   & 16.4   & \cellcolor{green!4}{0.445} 
                      & \cellcolor{green!16}{1.58} \\
    Llama-3.2-8B      & 37.6   & 37.6   & \cellcolor{red!1}{-0.040} 
                      & 67.0   & 69.2   & \cellcolor{red!22}{-2.20} 
                      & 6.67   & 0.00   & \cellcolor{green!67}{6.67} 
                      & 7.50   & 7.50   & \cellcolor{white}{0.00} 
                      & 9.19   & 11.0   & \cellcolor{red!18}{-1.78} 
                      & \cellcolor{green!5}{0.530} \\
    Llama-3.2-70B     & 74.5   & 72.2   & \cellcolor{green!23}{2.28} 
                      & 92.0   & 92.2   & \cellcolor{red!2}{-0.152} 
                      & 16.7   & 16.7   & \cellcolor{white}{0.00} 
                      & 67.5   & 50.0   & \cellcolor{green!100}{17.5} 
                      & 37.3   & 35.7   & \cellcolor{green!16}{1.63} 
                      & \cellcolor{green!43}{4.25} \\
    \midrule
    Qwen2.5-0.5B      & 30.0   & 31.0   & \cellcolor{red!9}{-0.920} 
                      & 43.1   & 45.4   & \cellcolor{red!24}{-2.35} 
                      & 0.00   & 0.00   & \cellcolor{white}{0.00} 
                      & 5.00   & 17.5   & \cellcolor{red!100}{-12.5} 
                      & 6.52   & 8.30   & \cellcolor{red!18}{-1.78} 
                      & \cellcolor{red!35}{-3.51} \\
    Qwen2.5-1.5B      & 50.3   & 50.7   & \cellcolor{red!4}{-0.440} 
                      & 70.6   & 71.0   & \cellcolor{red!5}{-0.455} 
                      & 0.00   & 3.33   & \cellcolor{red!33}{-3.33} 
                      & 22.5   & 20.0   & \cellcolor{green!25}{2.50} 
                      & 17.8   & 20.0   & \cellcolor{red!22}{-2.22} 
                      & \cellcolor{red!8}{-0.790} \\
    Qwen2.5-3B        & 57.5   & 60.3   & \cellcolor{red!28}{-2.82} 
                      & 79.9   & 79.5   & \cellcolor{green!4}{0.379} 
                      & 0.00   & 3.33   & \cellcolor{red!33}{-3.33} 
                      & 35.0   & 27.5   & \cellcolor{green!75}{7.50} 
                      & 25.9   & 26.4   & \cellcolor{red!4}{-0.444} 
                      & \cellcolor{green!3}{0.256} \\
    Qwen2.5-7B        & 71.3   & 63.6   & \cellcolor{green!77}{7.66} 
                      & 87.8   & 84.1   & \cellcolor{green!37}{3.72} 
                      & 6.67   & 0.00   & \cellcolor{green!67}{6.67} 
                      & 40.0   & 35.0   & \cellcolor{green!50}{5.00} 
                      & 38.8   & 29.0   & \cellcolor{green!98}{9.78} 
                      & \cellcolor{green!66}{6.56} \\
    Qwen2.5-14B       & 76.4   & 72.8   & \cellcolor{green!37}{3.66} 
                      & 93.1   & 89.6   & \cellcolor{green!35}{3.49} 
                      & 6.67   & 3.33   & \cellcolor{green!33}{3.33} 
                      & 47.5   & 45.0   & \cellcolor{green!25}{2.50} 
                      & 41.0   & 39.0   & \cellcolor{green!21}{2.07} 
                      & \cellcolor{green!30}{3.01} \\
    Qwen2.5-32B       & 80.5   & 76.8   & \cellcolor{green!37}{3.72} 
                      & 92.2   & 92.7   & \cellcolor{red!5}{-0.531} 
                      & 20.0   & 3.33   & \cellcolor{green!100}{16.7} 
                      & 57.5   & 50.0   & \cellcolor{green!75}{7.50} 
                      & 47.4   & 42.4   & \cellcolor{green!50}{5.04} 
                      & \cellcolor{green!65}{6.48} \\
    \bottomrule
  \end{tabular}
  }

\caption{This table summarizes the performance of models in Llama and Qwen families fine-tuned with large teacher CoT and small teacher CoT when evaluated on MATH, GSM8K, AIME, AMC, and OlympiadBench. \texttt{Qwen-2.5-72B-Instruct} is chosen as the large teacher while \texttt{Qwen-2.5-3B-Instruct} is chosen as the small teacher. We observe that small student models may experience degraded performance when distilled from a large teacher compared to a small teacher, whereas larger student models benefit more from the distilling a large teacher.}
\label{tab:lmp-full_comparison}
\end{table*}

%% file: tables/long_CoT_full_table.tex
\begin{table*}[htbp]
  \centering
  \resizebox{1\textwidth}{!}{%
  \begin{tabular}{l*{5}{ccc}c}
    \toprule
    & \multicolumn{3}{c}{MATH} 
    & \multicolumn{3}{c}{GSM8K} 
    & \multicolumn{3}{c}{AIME} 
    & \multicolumn{3}{c}{AMC} 
    & \multicolumn{3}{c}{Olympiad} 
    & \multicolumn{1}{c}{Average $\Delta_{\rm Long}$} \\
    \cmidrule(lr){2-4} \cmidrule(lr){5-7} \cmidrule(lr){8-10} \cmidrule(lr){11-13} \cmidrule(lr){14-16}
    Model 
    & $P_{\rm Long}$ & $P_{\rm Short}$ & $\Delta_{\rm Long}$ 
    & $P_{\rm Long}$ & $P_{\rm Short}$ & $\Delta_{\rm Long}$ 
    & $P_{\rm Long}$ & $P_{\rm Short}$ & $\Delta_{\rm Long}$ 
    & $P_{\rm Long}$ & $P_{\rm Short}$ & $\Delta_{\rm Long}$ 
    & $P_{\rm Long}$ & $P_{\rm Short}$ & $\Delta_{\rm Long}$ 
    & \\ 
    \midrule
    Llama-3.2-1B  & 28.6  & 33.4  & \cellcolor{red!32} -4.78  & 42.3  & 49.2  & \cellcolor{red!46} -6.90  & 0.00  & 0.00  & 0.00  & 2.50  & 7.50  & \cellcolor{red!33} -5.00  & 5.48  & 7.40  & \cellcolor{red!13} -1.92  & \cellcolor{red!25} -3.72 \\
    Llama-3.2-3B  & 48.7  & 50.9  & \cellcolor{red!14} -2.14  & 75.1  & 77.5  & \cellcolor{red!16} -2.42  & 3.33  & 3.33  & \cellcolor{white} 0.00  & 17.5  & 15.0  & \cellcolor{green!17} 2.50  & 17.6  & 18.7  & \cellcolor{red!7} -1.04  & \cellcolor{red!4} -0.619 \\
    Llama-3.1-8B  & 50.0  & 44.6  & \cellcolor{green!36} 5.36   & 81.4  & 75.5  & \cellcolor{green!39} 5.84   & 0.00  & 0.00  & \cellcolor{white} 0.00   & 27.5  & 22.5  & \cellcolor{green!33} 5.00   & 17.3  & 14.8  & \cellcolor{green!17} 2.52   & \cellcolor{green!25} 3.74 \\
    Llama-3.3-70B & 75.3  & 74.9  & \cellcolor{green!2} 0.340   & 92.7  & 91.2  & \cellcolor{green!10} 1.44   & 26.7  & 13.3  & \cellcolor{green!89} 13.3   & 55.0  & 52.5  & \cellcolor{green!17} 2.50   & 41.3  & 39.7  & \cellcolor{green!11} 1.63   & \cellcolor{green!26} 3.85 \\
    \midrule
    Qwen2.5-0.5B  & 23.0  & 31.5  & \cellcolor{red!56} -8.44  & 39.5  & 45.3  & \cellcolor{red!39} -5.84  & 0.00  & 0.00  & \cellcolor{white} 0.00  & 7.50  & 15.0  & \cellcolor{red!50} -7.50  & 4.00  & 5.93  & \cellcolor{red!13} -1.93  & \cellcolor{red!32} -4.74 \\
    Qwen2.5-1.5B  & 41.6  & 52.3  & \cellcolor{red!71} -10.7  & 63.8  & 71.7  & \cellcolor{red!53} -7.89  & 0.00  & 0.00  & \cellcolor{white} 0.00  & 17.5  & 27.5  & \cellcolor{red!67} -10.0  & 12.3  & 19.4  & \cellcolor{red!47} -7.11  & \cellcolor{red!48} -7.13 \\
    Qwen2.5-3B   & 56.2  & 61.0  & \cellcolor{red!32} -4.84  & 80.0  & 82.0  & \cellcolor{red!13} -1.98  & 3.33  & 10.0  & \cellcolor{red!44} -6.67  & 37.5  & 37.5  & \cellcolor{white} 0.00  & 24.4  & 26.4  & \cellcolor{red!13} -1.93  & \cellcolor{red!21} -3.08 \\
    Qwen2.5-7B   & 68.2  & 67.8  & \cellcolor{green!3} 0.460    & 86.2  & 85.7  & \cellcolor{green!4} 0.560    & 13.3  & 6.67  & \cellcolor{green!44} 6.67    & 40.0  & 40.0  & \cellcolor{white} 0.00   & 36.6  & 35.7  & \cellcolor{green!6} 0.889    & \cellcolor{green!11} 1.72 \\
    Qwen2.5-14B  & 78.3  & 76.2  & \cellcolor{green!14} 2.04   & 93.3  & 92.5  & \cellcolor{green!5} 0.760    & 20.0  & 6.67  & \cellcolor{green!89} 13.3   & 60.0  & 55.0  & \cellcolor{green!33} 5.00   & 44.4  & 40.9  & \cellcolor{green!24} 3.56   & \cellcolor{green!33} 4.94 \\
    Qwen2.5-32B  & 84.8  & 82.3  & \cellcolor{green!16} 2.44   & 94.9  & 94.3  & \cellcolor{green!4} 0.610    & 40.0  & 10.0  & \cellcolor{green!100} 30.0  & 85.0  & 62.5  & \cellcolor{green!100} 22.5  & 60.4  & 47.3  & \cellcolor{green!88} 13.2  & \cellcolor{green!91} 13.7 \\
    \bottomrule
  \end{tabular}
  }

\caption{This table summarizes the performance of models in Llama and Qwen families fine-tuned with long CoT and short CoT data. They are evaluated on MATH, GSM8K, AIME, AMC, and OlympiadBench. \texttt{QwQ-32B-Preview} is chosen to generate long CoT and awhile \texttt{Qwen-2.5-32B-Instruct} is chosen to generate short CoT. We observe that small student models tend to benefit more from short CoT, while large student models gain greater advantages from long CoT.}
\label{tab:full_performance_lg}
\end{table*}

%% file: tables/lmp_more_teachers.tex
\begin{table*}[htbp]
  \centering
  \resizebox{1\textwidth}{!}{%
  \begin{tabular}{l *{6}{c} *{6}{c}}
    \toprule
    & \multicolumn{6}{c}{Gemma2-9B vs Gemma2-27B} & \multicolumn{6}{c}{Llama3.1-8B vs Llama3.1-70B} \\
    \cmidrule(lr){2-7} \cmidrule(lr){8-13}
    Model & MATH & AMC & Olympiad & AIME & GSM8k & Average & MATH & AMC & Olympiad & AIME & GSM8k & Average \\
    \midrule
    Llama3.2-1B   & \cellcolor{white}{-1.42}  & \cellcolor{white}{-7.50}  & \cellcolor{white}{0.00}   & \cellcolor{white}{0.00}   & \cellcolor{white}{-0.227}  & \cellcolor{red!61}{-1.83}   & \cellcolor{white}{-1.42}  & \cellcolor{white}{-5.00}  & \cellcolor{white}{-0.296} & \cellcolor{white}{3.33}  & \cellcolor{white}{0.152}  & \cellcolor{red!22}{-0.646} \\
    Llama3.2-3B   & \cellcolor{white}{2.08}   & \cellcolor{white}{-7.50}  & \cellcolor{white}{-0.888} & \cellcolor{white}{0.00}   & \cellcolor{white}{1.67}   & \cellcolor{red!31}{-0.928}  & \cellcolor{white}{-0.14}  & \cellcolor{white}{10.0}   & \cellcolor{white}{-0.593} & \cellcolor{white}{3.33}  & \cellcolor{white}{1.06}   & \cellcolor{green!91}{2.73} \\
    Llama3.1-8B   & \cellcolor{white}{0.56}   & \cellcolor{white}{0.00}   & \cellcolor{white}{0.078} & \cellcolor{white}{0.00}   & \cellcolor{white}{-0.516}  & \cellcolor{green!5}{0.0243}  & \cellcolor{white}{-2.18}  & \cellcolor{white}{7.50}   & \cellcolor{white}{2.67}  & \cellcolor{white}{0.00}  & \cellcolor{white}{-1.29}  & \cellcolor{green!45}{1.34} \\
    Llama3.1-70B  & \cellcolor{white}{0.02}   & \cellcolor{white}{7.50}   & \cellcolor{white}{-0.741} & \cellcolor{white}{10.0}   & \cellcolor{white}{0.152}   & \cellcolor{green!100}{3.39}  & \cellcolor{white}{2.72}   & \cellcolor{white}{17.5}   & \cellcolor{white}{5.48}  & \cellcolor{white}{6.67}  & \cellcolor{white}{0.986}   & \cellcolor{green!100}{6.67} \\
\midrule
    Qwen2.5-0.5B  & \cellcolor{white}{-4.56}  & \cellcolor{white}{0.00}   & \cellcolor{white}{0.741} & \cellcolor{white}{0.00}   & \cellcolor{white}{0.592}   & \cellcolor{red!22}{-0.645}  & \cellcolor{white}{-1.88}  & \cellcolor{white}{0.00}   & \cellcolor{white}{0.185} & \cellcolor{white}{0.00}  & \cellcolor{white}{-1.74}  & \cellcolor{red!23}{-0.688} \\
    Qwen2.5-1.5B  & \cellcolor{white}{-1.20}  & \cellcolor{white}{2.50}   & \cellcolor{white}{-1.19} & \cellcolor{white}{0.00}   & \cellcolor{white}{-0.986}  & \cellcolor{red!6}{-0.174}  & \cellcolor{white}{-1.48}  & \cellcolor{white}{5.00}   & \cellcolor{white}{-0.148} & \cellcolor{white}{3.33}  & \cellcolor{white}{-1.14}  & \cellcolor{green!37}{1.11} \\
    Qwen2.5-3B    & \cellcolor{white}{0.44}   & \cellcolor{white}{5.00}   & \cellcolor{white}{1.78}  & \cellcolor{white}{0.00}   & \cellcolor{white}{-0.758}  & \cellcolor{green!43}{1.29}   & \cellcolor{white}{-1.26}  & \cellcolor{white}{5.00}   & \cellcolor{white}{-0.741} & \cellcolor{white}{-3.33} & \cellcolor{white}{-1.29}  & \cellcolor{red!11}{-0.325} \\
    Qwen2.5-7B    & \cellcolor{white}{0.22}   & \cellcolor{white}{5.00}   & \cellcolor{white}{1.04}  & \cellcolor{white}{-3.33}  & \cellcolor{white}{3.94}   & \cellcolor{green!46}{1.37}   & \cellcolor{white}{3.68}   & \cellcolor{white}{20.0}   & \cellcolor{white}{4.15}  & \cellcolor{white}{3.33}  & \cellcolor{white}{2.81}   & \cellcolor{green!100}{6.79} \\
    Qwen2.5-14B   & \cellcolor{white}{1.32}   & \cellcolor{white}{2.50}   & \cellcolor{white}{-0.148} & \cellcolor{white}{0.00}   & \cellcolor{white}{-0.986}  & \cellcolor{green!18}{0.537}  & \cellcolor{white}{2.18}   & \cellcolor{white}{0.00}   & \cellcolor{white}{0.445}  & \cellcolor{white}{3.33}  & \cellcolor{white}{-0.303}  & \cellcolor{green!38}{1.13} \\
    Qwen2.5-32B   & \cellcolor{white}{0.10}   & \cellcolor{white}{2.50}   & \cellcolor{white}{1.48}  & \cellcolor{white}{3.44}   & \cellcolor{white}{1.36}   & \cellcolor{green!59}{1.78}  & \cellcolor{white}{2.72}   & \cellcolor{white}{-2.50}  & \cellcolor{white}{5.63}  & \cellcolor{white}{3.33}  & \cellcolor{white}{0.834}   & \cellcolor{green!67}{2.00} \\
    \bottomrule
  \end{tabular}
  }
  \caption{This table presents the performance of student models distilled from different teacher models, including \texttt{Gemma-27B-it} vs \texttt{Gemma-9B-it} and \texttt{Llama3.1-72B-Instruct} vs \texttt{Llama3.1-8B-Instruct}. We observe that small student models may experience degraded performance when distilled from a large teacher compared to a small teacher, whereas larger student models benefit more from the distilling a large teacher.}

  \label{tab:lmp_comparison2}
\end{table*}

%% file: appendix/examples_of_different_CoT.tex
\section{Examples of Speaking Style Shift}
\label{Examples of Speaking Style Shift}
\begin{figure*}[!t]
    \centering
    \includegraphics[width=1\textwidth]{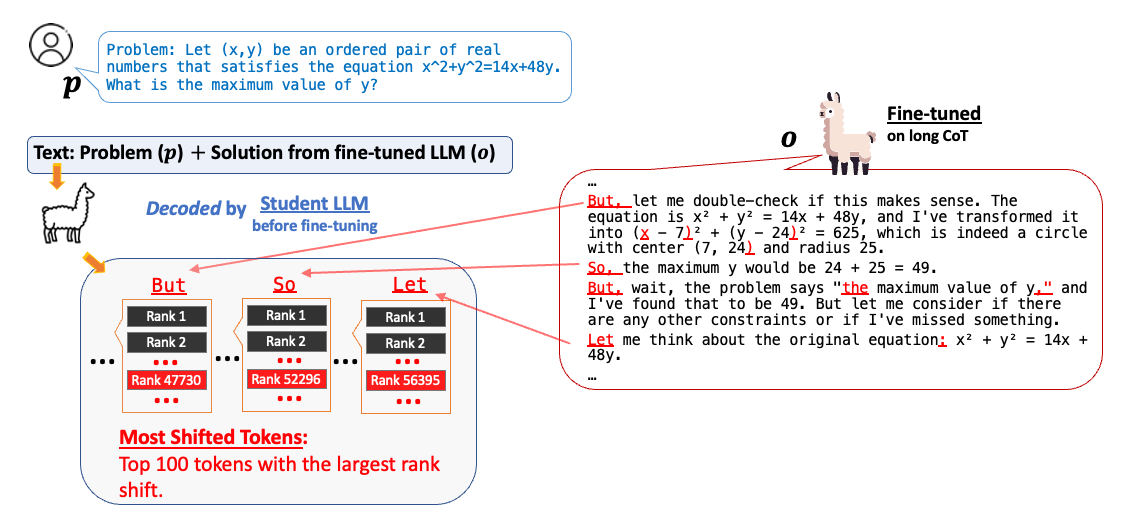}
    \caption{The process of calculating most shifted tokens. We decode each token generated by the fine-tuned LLM in the student model before fine-tuning. Then we calculate the rank shift in the student model for each token generated by the fine-tuned model. We annotate the tokens that exhibit the largest rank shifts as the most shifted tokens. We found that these tokens are predominantly associated with expressive and stylistic elements, such as “But” and “Let”.}
    \label{fig:speaking_style}
\end{figure*}

We adopt the method from  \citep{lin2023unlockingspellbasellms} to evaluate the most shifted tokens after fine-tuning on long CoT and Large teacher CoT data. Figure \ref{fig:speaking_style} shows the calculation process. This allows us to compare the token distribution shifts induced by the fine-tuning process. We annotate the tokens that exhibit the largest rank shifts as the most shifted tokens. We choose \texttt{Qwen2.5-3B-Instruct} as the student model. We put the results of most shifted tokens after fine-tuning on long CoT data in Figure \ref{fig: most_shifted_1} and \ref{fig: most_shifted_2}. The results of most shifted tokens after fine-tuning on large teacher CoT data are shown in Figure \ref{fig: most_shifted_3}. Our analysis reveals that these tokens are predominantly associated with expressive and stylistic elements, such as “wait”, “But”, and “Let”.

\begin{figure*}[htbp]
    \centering
\begin{tcolorbox}[title=Problem, promptstyle]
\lstset{
    basicstyle=\normalfont\sffamily\footnotesize,
    breaklines=true,
    frame=none,
    columns=fullflexible,
}
Let $(x,y)$ be an ordered pair of real numbers that satisfies the equation $x^2+y^2=14x+48y$. What is the maximum value of $y$?
\end{tcolorbox}
\end{figure*}

\begin{figure*}[htbp]
    \centering
\begin{tcolorbox}[title=Example of \textcolor{red}{most shifted tokens} after fine-tuning on long CoT data. (1/2), promptstyle]
\lstset{
    basicstyle=\normalfont\sffamily\footnotesize,
    breaklines=true,
    frame=none,
    columns=fullflexible,
}
So I've got this math problem here\textcolor{red}{.} It says\textcolor{red}{:} ``\textcolor{red}{Let} (\textcolor{red}{x}\textcolor{red}{,} y\textcolor{red}{)} be an ordered pair of real numbers that satisfies the equation \(x^2 + y^2 = 14x + 48y\)\textcolor{red}{.} What is the maximum value of \(y\)\textcolor{red}{?} Okay\textcolor{red}{,} let's see how to approach this.

\textcolor{red}{First}\textcolor{red}{,} I need to understand what the equation is telling me\textcolor{red}{.} It's an equation involving \(x\) and \(y\)\textcolor{red}{,} and it's set equal to zero\textcolor{red}{:} 
\[
x^2 + y^2 - 14x - 48y = 0.
\]
I need to find the maximum value of \(y\) for real numbers \(x\) and \(y\) that satisfy this equation.

Hmm\textcolor{red}{,} maybe I can rearr\textcolor{red}{ange} this equation to make it easier to work with\textcolor{red}{.} Let's see:
\[
x^2 - 14x + y^2 - 48y = 0.
\]
I can complete the square for both \(x\) and \(y\) terms.

\textcolor{red}{For} \(x\)\textcolor{red}{:} \(x^2 - 14x\)\textcolor{red}{,} to complete the square\textcolor{red}{,} I take half of \(-14\), which is \(-7\), and square it to get \(49\)\textcolor{red}{.} So\textcolor{red}{,} \(x^2 - 14x\) becomes \((x - 7\textcolor{red}{)}^2 - 49\).

\textcolor{red}{For} \(y\): \(y^2 - 48y\), half of \(-48\) is \(-24\), squared is \(576\)\textcolor{red}{.} So, \(y^2 - 48y\) becomes \((y - 24\textcolor{red}{)}^2 - 576\).

\textcolor{red}{Now}\textcolor{red}{,} pl\textcolor{red}{ugging} these back into the equation:
\[
(x - 7\textcolor{red}{)}^2 - 49 + (y - 24\textcolor{red}{)}^2 - 576 = 0.
\]
S\textcolor{red}{implify}:
\[
(x - 7\textcolor{red}{)}^2 + (y - 24)^2 - 625 = 0,
\]
\[
\text{So, } (x - 7\textcolor{red}{)}^2 + (y - 24)^2 = 625.
\]

\textcolor{red}{This} looks like the equation of a circle with center at \((7, 24\textcolor{red}{)}\) and radius \(25\).

Wait\textcolor{red}{,} but the problem is asking for the maximum value of \(y\)\textcolor{red}{,} and I have a circle equation\textcolor{red}{.} \textcolor{red}{So}\textcolor{red}{,} the maximum \(y\) would be the \(y\)-coordinate of the highest point on the circle\textcolor{red}{,} which is the center plus the radius.

Wait\textcolor{red}{,} but the center is at \((7\textcolor{red}{,} 24)\), and the radius is \(25\)\textcolor{red}{,} so the highest point would be at \((7\textcolor{red}{,} 24 + 25) = (7\textcolor{red}{,} 49)\).

\textcolor{red}{But}\textcolor{red}{,} let me double-check if this makes sense\textcolor{red}{.} The equation is 
\[
x^2 + y^2 = 14x + 48y,
\]
and I've transformed it into 
\[
(\textcolor{red}{x} - 7\textcolor{red}{)}^2 + (y - 24)^2 = 625,
\]
which is indeed a circle with center \((7, 24\textcolor{red}{)}\) and radius \(25\).

\textcolor{red}{So}\textcolor{red}{,} the maximum \(y\) would be \(24 + 25 = 49\).

\textcolor{red}{But}\textcolor{red}{,} wait\textcolor{red}{,} the problem says ``\textcolor{red}{the} maximum value of \(y\)\textcolor{red}{,}'' and I've found that to be \(49\)\textcolor{red}{.} But let me consider if there are any other constraints or if I've missed something.

\textcolor{red}{Let} me think about the original equation:
\[
x^2 + y^2 = 14x + 48y.
\]
I can rearr\textcolor{red}{ange} it to:
\[
x^2 - 14x + y^2 - 48y = 0.
\]

\end{tcolorbox}
\caption{Example of \textcolor{red}{most shifted tokens} after fine-tuning on long CoT data. All the most shifted tokens are labeled as red.}
    \label{fig: most_shifted_1}
\end{figure*}

\begin{figure*}[htbp]
    \centering
\begin{tcolorbox}[title=Example of \textcolor{red}{most shifted tokens} after fine-tuning on long CoT data (2/2), promptstyle]
\lstset{
    basicstyle=\normalfont\sffamily\footnotesize,
    breaklines=true,
    frame=none,
    columns=fullflexible,
}
\textcolor{red}{Then}\textcolor{red}{,} completing the square:
\[
(x - 7\textcolor{red}{)}^2 - 49 + (y - 24)^2 - 576 = 0,
\]
\[
(x - 7\textcolor{red}{)}^2 + (y - 24)^2 = 625.
\]
Yes\textcolor{red}{,} that's a circle with center \((7, 24\textcolor{red}{)}\) and radius \(25\).

\textcolor{red}{So}\textcolor{red}{,} the maximum \(y\) is indeed \(24 + 25 = 49\).

\textcolor{red}{But}\textcolor{red}{,} just to be thorough\textcolor{red}{,} let's consider if there are any points on the circle where \(y\) is greater than \(49\)\textcolor{red}{.} The center is at \((7, 24\textcolor{red}{)}\), and the radius is \(25\)\textcolor{red}{,} so the highest point is at \((7, 49\textcolor{red}{)}\), and the lowest point would be at \((7\textcolor{red}{,} -1)\).

\textcolor{red}{But}\textcolor{red}{,} the problem is asking for the maximum value of \(y\)\textcolor{red}{,} and I've found that to be \(49\).

Therefore\textcolor{red}{,} the final answer is \(49\).

\bigskip

\textbf{\textcolor{red}{Final} Answer:}

\[
\boxed{49\textcolor{red}{}}
\]
\end{tcolorbox}
\caption{\textcolor{red}{Most shifted tokens} after fine-tuning on long CoT data (2/2). All the most shifted tokens are labeled as red.}
    \label{fig: most_shifted_2}
\end{figure*}

\begin{figure*}[htbp]
    \centering
\begin{tcolorbox}[title=Example of \textcolor{red}{Most Shifted Tokens} after fine-tuning on large model CoT, promptstyle]
\lstset{
    basicstyle=\normalfont\sffamily\footnotesize,
    breaklines=true,
    frame=none,
    columns=fullflexible,
}

\textcolor{red}{To} \textcolor{red}{solve} \textcolor{red}{the} \textcolor{red}{problem}\textcolor{red}{,} \textcolor{red}{ we} \textcolor{red}{ start} \textcolor{red}{ with} \textcolor{red}{ the} \textcolor{red}{ given} equation\textcolor{red}{:}

\[
x^2 + y^2\textcolor{red}{ =} 14x + 48y
\]

\textcolor{red}{First}\textcolor{red}{,} we rearr\textcolor{red}{ange} the equation to bring all terms to one\textcolor{red}{ side}\textcolor{red}{:}

\[
x^2 - 14\textcolor{red}{x} + y^2 - 48\textcolor{red}{y} = 0
\]

\textcolor{red}{Next}\textcolor{red}{,} we complete the square for both \(x\) and \(\textcolor{red}{y}\).

\textcolor{red}{For} \(x\):
\[
x^2 - 14x
\]
\textcolor{red}{To} complete the square\textcolor{red}{,} we add and subtract \(\left(\frac{14}{2}\right)^2 = 49\):
\[
x^2 - 14x + 49 - 49
\]
\[
(\textcolor{red}{x} - 7)^2 - 49
\]

\textcolor{red}{For} \(y\):
\[
y^2 - 48y
\]
\textcolor{red}{To} complete the square, we add and subtract \(\left(\frac{48}{2}\right)^2 = 576\):
\[
y^2 - 48y + 576 - 576
\]
\[
(\textcolor{red}{y} - 24)^2 - 576
\]

\textcolor{red}{Now}\textcolor{red}{,} substitute these completed squares back into the equation:
\[
(\textcolor{red}{x} - 7)^2 - 49 + (y - 24)^2 - 576 = 0
\]
\[
(\textcolor{red}{x} - 7)^2 + (y - 24)^2 - 625 = 0
\]
\[
(\textcolor{red}{x} - 7)^2 + (y - 24)^2 = 625
\]

\textcolor{red}{This} is the equation of a circle with center \((7, 24)\) and radius \(25\).

\textcolor{red}{To} find the maximum value of \(y\), we need to find the highest point on the circle. \textcolor{red}{The} highest point on the circle is directly above the center, which is at \(y = 24 + 25 = 49\).

\textcolor{red}{Thus}\textcolor{red}{,} the maximum value of \(y\) is:
\[
\textcolor{red}{y} = 49
\]

\textcolor{red}{Final} \textcolor{red}{ Answer}\textcolor{red}{:} \(\boxed{49}\)

\end{tcolorbox}
\caption{Example of \textcolor{red}{most shifted tokens} after fine-tuning on large teacher CoT. All the most shifted tokens are labeled as red.}
    \label{fig: most_shifted_3}
\end{figure*}

\section{Examples of Various CoT Data}\label{app:example}
This section demonstrates examples of different CoT data including long CoT, short CoT, large teacher CoT and small teacher CoT. See the following examples.

\begin{figure*}[htbp]
    \centering
\begin{tcolorbox}[title=Problem, promptstyle]
\lstset{
    basicstyle=\normalfont\sffamily\footnotesize,
    breaklines=true,
    frame=none,
    columns=fullflexible,
}
What is the sum of the positive odd divisors of $60$?
\end{tcolorbox}
    
\end{figure*}

\begin{figure*}[htbp]
    \centering
\begin{tcolorbox}[title=Long CoT Example, promptstyle]
\lstset{
    basicstyle=\normalfont\sffamily\footnotesize,
    breaklines=true,
    frame=none,
    columns=fullflexible,
}
To find the sum of the positive odd divisors of 60, I first need to understand what the divisors of 60 are and then identify which of those are odd.

First, let's find all the positive divisors of 60. To do this, I can start by factoring 60 into its prime factors.

\[
60 = 2^2 \times 3^1 \times 5^1
\]

Now, to find all the positive divisors, I can take all possible products of these prime factors raised to their possible exponents. For 2, the exponents can be \(0, 1,\) or \(2\); for 3, \(0\) or \(1\); and for 5, \(0\) or \(1\). Listing them out:

\[
\begin{aligned}
&1 = 2^0 \times 3^0 \times 5^0, \\
&2 = 2^1 \times 3^0 \times 5^0, \\
&4 = 2^2 \times 3^0 \times 5^0, \\
&3 = 2^0 \times 3^1 \times 5^0, \\
&6 = 2^1 \times 3^1 \times 5^0, \\
&12 = 2^2 \times 3^1 \times 5^0, \\
&5 = 2^0 \times 3^0 \times 5^1, \\
&10 = 2^1 \times 3^0 \times 5^1, \\
&20 = 2^2 \times 3^0 \times 5^1, \\
&15 = 2^0 \times 3^1 \times 5^1, \\
&30 = 2^1 \times 3^1 \times 5^1, \\
&60 = 2^2 \times 3^1 \times 5^1.
\end{aligned}
\]

So, the positive divisors of 60 are:
\[
1, 2, 3, 4, 5, 6, 10, 12, 15, 20, 30, 60.
\]

Now, I need to identify which of these are odd. Odd numbers are not divisible by 2, so I can eliminate any divisors that are even.

The odd divisors are:
\[
1, 3, 5, 15.
\]

Now, I need to find the sum of these odd divisors:
$1 + 3 + 5 + 15 = 24$.
Alternatively, using the prime factorization approach: Since the odd divisors are those that do not include the factor \(2\), I can ignore the 2's in the prime factorization. The remaining odd part is:

\[
3^1 \times 5^1.
\]

The sum of the divisors of this odd part is:

\[
(1 + 3)(1 + 5) = 4 \times 6 = 24.
\]

This matches the previous result.

Therefore, for \( 60 = 2^2 \times 3^1 \times 5^1 \), the sum of the odd divisors is:

\[
\boxed{24}.
\]
\end{tcolorbox}
\caption{Long CoT Example}
    \label{fig: long cot example}
\end{figure*}

\begin{figure*}[htbp]
    \centering
\begin{tcolorbox}[title=Short CoT Example, promptstyle]
\lstset{
    basicstyle=\normalfont\sffamily\footnotesize,
    breaklines=true,
    frame=none,
    columns=fullflexible,
}

To find the sum of the positive odd divisors of $60$, we first need to factorize $60$ into its prime factors. The prime factorization of $60$ is:

\[
60 = 2^2 \times 3 \times 5
\]

Since we are only interested in the odd divisors, we can ignore the factor of $2$. Therefore, we only consider the factors $3$ and $5$. The odd divisors of $60$ are the divisors of $3 \times 5 = 15$. The divisors of $15$ are \(1, 3, 5,\) and \(15\).

Next, we sum these divisors:

\[
1 + 3 + 5 + 15 = 24
\]

Therefore, the sum of the positive odd divisors of $60$ is:

\[
\boxed{24}
\]

\end{tcolorbox}
    \label{fig: short cot example}
    \caption{Short CoT example}
\end{figure*}

\begin{figure*}[htbp]
    \centering
\begin{tcolorbox}[title=Problem, promptstyle]
\lstset{
    basicstyle=\normalfont\sffamily\footnotesize,
    breaklines=true,
    frame=none,
    columns=fullflexible,
}
Find the greatest \( a \) such that
\[
\frac{7\sqrt{(2a)^2+(1)^2}-4a^2-1}{\sqrt{1+4a^2}+3}=2.
\]
\end{tcolorbox}
\end{figure*}

\begin{figure*}[htbp]
    \centering
\begin{tcolorbox}[title=Large Teacher CoT Example (1/2), promptstyle]
\lstset{
    basicstyle=\normalfont\sffamily\footnotesize,
    breaklines=true,
    frame=none,
    columns=fullflexible,
}

To solve the equation \(\frac{7\sqrt{(2a)^2 + 1^2} - 4a^2 - 1}{\sqrt{1 + 4a^2} + 3} = 2\), we start by simplifying the expression inside the square root in the numerator. The equation becomes:

\[
\frac{7\sqrt{4a^2 + 1} - 4a^2 - 1}{\sqrt{1 + 4a^2} + 3} = 2
\]

Let's denote \(x = \sqrt{1 + 4a^2}\). Then the equation transforms into:

\[
\frac{7x - 4a^2 - 1}{x + 3} = 2
\]

Next, we clear the fraction by multiplying both sides by \(x + 3\):

\[
7x - 4a^2 - 1 = 2(x + 3)
\]

Expanding and simplifying the right side, we get:

\[
7x - 4a^2 - 1 = 2x + 6
\]

Rearranging terms to isolate \(x\) and \(a^2\), we have:

\[
7x - 2x - 4a^2 - 1 - 6 = 0 \implies 5x - 4a^2 - 7 = 0 \implies 5x = 4a^2 + 7 \implies x = \frac{4a^2 + 7}{5}
\]

Recall that \(x = \sqrt{1 + 4a^2}\). Therefore, we have:

\[
\sqrt{1 + 4a^2} = \frac{4a^2 + 7}{5}
\]

To eliminate the square root, we square both sides:

\[
1 + 4a^2 = \left(\frac{4a^2 + 7}{5}\right)^2
\]

Expanding the right side, we get:

\[
1 + 4a^2 = \frac{(4a^2 + 7)^2}{25} = \frac{16a^4 + 56a^2 + 49}{25}
\]

Multiplying both sides by 25 to clear the denominator, we obtain:

\[
25 + 100a^2 = 16a^4 + 56a^2 + 49
\]

\end{tcolorbox}
\end{figure*}

\begin{figure*}[htbp]
    \centering
\begin{tcolorbox}[title=Large Teacher CoT Example (2/2), promptstyle]
\lstset{
    basicstyle=\normalfont\sffamily\footnotesize,
    breaklines=true,
    frame=none,
    columns=fullflexible,
}
Rearranging all terms to one side, we get a standard polynomial equation:

\[
16a^4 + 56a^2 + 49 - 100a^2 - 25 = 0 \implies 16a^4 - 44a^2 + 24 = 0
\]

Let \(y = a^2\). Then the equation becomes a quadratic in \(y\):

\[
16y^2 - 44y + 24 = 0
\]

We solve this quadratic equation using the quadratic formula \(y = \frac{-b \pm \sqrt{b^2 - 4ac}}{2a}\), where \(a = 16\), \(b = -44\), and \(c = 24\):

\[
y = \frac{44 \pm \sqrt{(-44)^2 - 4 \cdot 16 \cdot 24}}{2 \cdot 16} = \frac{44 \pm \sqrt{1936 - 1536}}{32} = \frac{44 \pm \sqrt{400}}{32} = \frac{44 \pm 20}{32}
\]

This gives us two solutions for \(y\):

\[
y = \frac{64}{32} = 2 \quad \text{and} \quad y = \frac{24}{32} = \frac{3}{4}
\]

Since \(y = a^2\), we have \(a^2 = 2\) or \(a^2 = \frac{3}{4}\). The corresponding values for \(a\) are \(a = \pm \sqrt{2}\) and \(a = \pm \frac{\sqrt{3}}{2}\). The greatest value of \(a\) is \(\sqrt{2}\).

Thus, the greatest \(a\) that satisfies the equation is:

\[
\boxed{\sqrt{2}}
\]
\end{tcolorbox}
\end{figure*}

\begin{figure*}[htbp]
    \centering
\begin{tcolorbox}[title=Small Teacher CoT Example (1/2), promptstyle]
\lstset{
    basicstyle=\normalfont\sffamily\footnotesize,
    breaklines=true,
    frame=none,
    columns=fullflexible,
}

To solve the equation \(\frac{7\sqrt{(2a)^2+(1)^2}-4a^2-1}{\sqrt{1+4a^2}+3}=2\), we start by simplifying the expression inside the square root and the equation itself.

First, simplify the expression under the square root in the numerator:
\[
(2a)^2 + 1^2 = 4a^2 + 1
\]
So the equation becomes:
\[
\frac{7\sqrt{4a^2 + 1} - 4a^2 - 1}{\sqrt{4a^2 + 1} + 3} = 2
\]

Next, let \( x = \sqrt{4a^2 + 1} \). Then the equation transforms into:
\[
\frac{7x - 4a^2 - 1}{x + 3} = 2
\]

Since \( x = \sqrt{4a^2 + 1} \), we can express \( 4a^2 \) in terms of \( x \):
\[
4a^2 = x^2 - 1
\]

Substitute \( 4a^2 = x^2 - 1 \) into the equation:
\[
\frac{7x - (x^2 - 1) - 1}{x + 3} = 2
\]
Simplify the numerator:
\[
7x - x^2 + 1 - 1 = 7x - x^2
\]
So the equation becomes:
\[
\frac{7x - x^2}{x + 3} = 2
\]

Multiply both sides by \( x + 3 \) to clear the denominator:
\[
7x - x^2 = 2(x + 3)
\]
Expand and simplify:
\[
7x - x^2 = 2x + 6
\]
Rearrange all terms to one side:
\[
-x^2 + 7x - 2x - 6 = 0
\]
\[
-x^2 + 5x - 6 = 0
\]
Multiply through by -1 to make the leading coefficient positive:
\[
x^2 - 5x + 6 = 0
\]

\end{tcolorbox}
    
\end{figure*}

\begin{figure*}[htbp]
    \centering
\begin{tcolorbox}[title=Small Teacher CoT Example (2/2), promptstyle]
\lstset{
    basicstyle=\normalfont\sffamily\footnotesize,
    breaklines=true,
    frame=none,
    columns=fullflexible,
}

Factor the quadratic equation:
\[
(x - 2)(x - 3) = 0
\]

So, the solutions are:
\[
x = 2 \quad \text{or} \quad x = 3
\]

Recall that \( x = \sqrt{4a^2 + 1} \). We need to check both solutions:

1. For \( x = 2 \):
\[
\sqrt{4a^2 + 1} = 2
\]
Square both sides:
\[
4a^2 + 1 = 4
\]
\[
4a^2 = 3
\]
\[
a^2 = \frac{3}{4}
\]
\[
a = \pm \frac{\sqrt{3}}{2}
\]

2. For \( x = 3 \):
\[
\sqrt{4a^2 + 1} = 3
\]
Square both sides:
\[
4a^2 + 1 = 9
\]
\[
4a^2 = 8
\]
\[
a^2 = 2
\]
\[
a = \pm \sqrt{2}
\]

The greatest value of \( a \) among the solutions is \( \sqrt{2} \).

Thus, the final answer is:
\[
\boxed{\sqrt{2}}
\]
\end{tcolorbox}
    
\end{figure*}

\begin{figure*}[htbp]
    \centering
\begin{tcolorbox}[title=Problem, promptstyle]
\lstset{
    basicstyle=\normalfont\sffamily\footnotesize,
    breaklines=true,
    frame=none,
    columns=fullflexible,
}
What is the greatest common divisor of 128, 144 and 480?
\end{tcolorbox}
    
\end{figure*}

\begin{figure*}[htbp]
    \centering
\begin{tcolorbox}[title=Large Teacher CoT Example, promptstyle]
\lstset{
    basicstyle=\normalfont\sffamily\footnotesize,
    breaklines=true,
    frame=none,
    columns=fullflexible,
}
To find the greatest common divisor (GCD) of 128, 144, and 480, we can use the prime factorization method.

\begin{enumerate}
    \item \textbf{Prime Factorization:}
    \begin{align*}
        128 &= 2^7, \\
        144 &= 2^4 \times 3^2, \\
        480 &= 2^5 \times 3 \times 5.
    \end{align*}

    \item \textbf{Identify the common prime factors:}\\
    The common prime factor among 128, 144, and 480 is \(2\).

    \item \textbf{Determine the lowest power of the common prime factor:}\\
    The lowest power of \(2\) in the factorizations is \(2^4\).

    \item \textbf{Calculate the GCD:}\\
    The GCD is:
    \[
    2^4 = 16.
    \]
\end{enumerate}

Therefore, the greatest common divisor of 128, 144, and 480 is:
\[
\boxed{16}
\]
\end{tcolorbox}
    
\end{figure*}

\begin{figure*}[htbp]
    \centering
\begin{tcolorbox}[title=Small Teacher CoT Example, promptstyle]
\lstset{
    basicstyle=\normalfont\sffamily\footnotesize,
    breaklines=true,
    frame=none,
    columns=fullflexible,
}

To find the greatest common divisor (GCD) of 128, 144, and 480, we start by finding the prime factorization of each number.

The prime factorization of 128 is:
\[
\begin{aligned}
128 \div 2 &= 64 \\
64 \div 2 &= 32 \\
32 \div 2 &= 16 \\
16 \div 2 &= 8 \\
8 \div 2 &= 4 \\
4 \div 2 &= 2 \\
2 \div 2 &= 1
\end{aligned}
\]
Thus, 
\[
128 = 2^7.
\]

The prime factorization of 144 is:
\[
\begin{aligned}
144 \div 2 &= 72 \\
72 \div 2 &= 36 \\
36 \div 2 &= 18 \\
18 \div 2 &= 9 \\
9 \div 3 &= 3 \\
3 \div 3 &= 1
\end{aligned}
\]
Thus,
\[
144 = 2^4 \times 3^2.
\]

The prime factorization of 480 is:
\[
\begin{aligned}
480 \div 2 &= 240 \\
240 \div 2 &= 120 \\
120 \div 2 &= 60 \\
60 \div 2 &= 30 \\
30 \div 2 &= 15 \\
15 \div 3 &= 5 \\
5 \div 5 &= 1
\end{aligned}
\]
Thus,
\[
480 = 2^5 \times 3 \times 5.
\]

Now, we identify the common prime factors. The factorizations are:
\[
128 = 2^7, \quad 144 = 2^4 \times 3^2, \quad 480 = 2^5 \times 3 \times 5.
\]
The only common prime factor among all three numbers is \(2\).

Next, we determine the smallest power of the common prime factor. The lowest power of \(2\) among these factorizations is \(2^4\).

Therefore, the GCD of 128, 144, and 480 is:
\[
\boxed{16}
\]

\end{tcolorbox}
    
\end{figure*}

\begin{figure*}[htbp]
    \centering
\begin{tcolorbox}[title=Problem, promptstyle]
\lstset{
    basicstyle=\normalfont\sffamily\footnotesize,
    breaklines=true,
    frame=none,
    columns=fullflexible,
}
Rounded to 2 decimal places, what is $\frac{7}{9}$?
\end{tcolorbox}
    
\end{figure*}

\begin{figure*}[htbp]
    \centering
\begin{tcolorbox}[title=Large Teacher CoT Example, promptstyle]
\lstset{
    basicstyle=\normalfont\sffamily\footnotesize,
    breaklines=true,
    frame=none,
    columns=fullflexible,
}
To solve the problem, we need to divide 7 by 9 and then round the result to two decimal places.

\[
\frac{7}{9} = 0.777\ldots
\]

Rounded to two decimal places, this becomes 0.78.

Final Answer: \(\boxed{0.78}\)
\end{tcolorbox}
    
\end{figure*}

\begin{figure*}[htbp]
    \centering
\begin{tcolorbox}[title=Small Teacher CoT Example, promptstyle]
\lstset{
    basicstyle=\normalfont\sffamily\footnotesize,
    breaklines=true,
    frame=none,
    columns=fullflexible,
}
To solve the problem of finding the value of \(\frac{7}{9}\) rounded to 2 decimal places, we can follow these steps:

\begin{enumerate}
    \item \textbf{Perform the division}: Calculate \(\frac{7}{9}\).
    \item \textbf{Convert to decimal}: Perform the division to get the decimal equivalent.
    \item \textbf{Round to 2 decimal places}: Round the decimal to 2 decimal places.
\end{enumerate}

Let's start with the division:

\[
\frac{7}{9} \approx 0.7777\ldots
\]

Next, we need to round this decimal to 2 decimal places. The digit in the third decimal place is 7, which is greater than or equal to 5. Therefore, we round the second decimal place up from 7 to 8.

So, \(0.7777\ldots\) rounded to 2 decimal places is \(0.78\).

The final answer is:

\[
\boxed{0.78}
\]
\end{tcolorbox}
    
\end{figure*}